\documentclass{article}

\usepackage{PRIMEarxiv}

\usepackage[utf8]{inputenc} % allow utf-8 input
\usepackage[T1]{fontenc}    % use 8-bit T1 fonts
\usepackage{hyperref}       % hyperlinks
\usepackage{url}            % simple URL typesetting
\usepackage{booktabs}       % professional-quality tables
\usepackage{amsfonts}       % blackboard math symbols
\usepackage{nicefrac}       % compact symbols for 1/2, etc.
\usepackage{microtype}      % microtypography
\usepackage{lipsum}
\usepackage{fancyhdr}       % header
\usepackage{graphicx}       % graphics
\graphicspath{{media/}}     % organize your images and other figures under media/ folder
\usepackage{caption} 
\usepackage{amsmath}
\usepackage{cleveref}
\usepackage{array}
\usepackage{pifont} 

\title{When Robots Do the Chores: \\
A Benchmark and Agent for \\
 Long-Horizon Household Task Execution} 

\author{
  \textbf{Zilin Zhu$^{1,2,3,*}$,
  Longteng Guo$^{1,*}$,
  Yanghong Mei$^{1,3,*}$,
  Bowen Pang$^{3}$,} \\
  \textbf{Zongxun Zhang$^{3}$, 
  Xingjian He$^{1}$,
  Ruyi Ji$^{1}$,
  Jing Liu$^{1,3,\dagger}$} \\
  $^{1}$Institute of Automation, Chinese Academy of Sciences, Beijing, China \\
  $^{2}$Zhongguancun Academy, Beijing, China \\
  $^{3}$University of Chinese Academy of Sciences, Beijing, China \\
  \texttt{smzzlcn@gmail.com} \\
  $^{*}$Equal contribution. \quad
  $^{\dagger}$Corresponding author.
}

\begin{document}
\maketitle

\begin{abstract}

Long-horizon household tasks demand robust high-level planning and sustained reasoning capabilities, which are largely overlooked by existing embodied AI benchmarks that emphasize short-horizon navigation or manipulation and rely on fixed task categories. We introduce LongAct, a benchmark designed to evaluate planning-level autonomy in long-horizon household tasks specified through free-form instructions. By abstracting away embodiment-specific low-level control, LongAct isolates high-level cognitive capabilities such as instruction understanding, dependency management, memory maintenance, and adaptive planning. We further propose HoloMind, a VLM-driven agent with a DAG-based long-horizon hierarchical planner, a Multimodal Spatial Memory for persistent world modeling, an Episodic Memory for experience reuse, and a global Critic for reflective supervision. Experiments with GPT-5 and Qwen3-VL models show that HoloMind substantially improves long-horizon performance while reducing reliance on model scale. Even top models achieve only 59\% goal completion  and 16\% full-task success, underscoring the difficulty of LongAct and the need for stronger long-horizon planning in embodied agents.


\end{abstract}    

\begin{figure*}[t]
    \centering
    \includegraphics[width=0.93\linewidth]{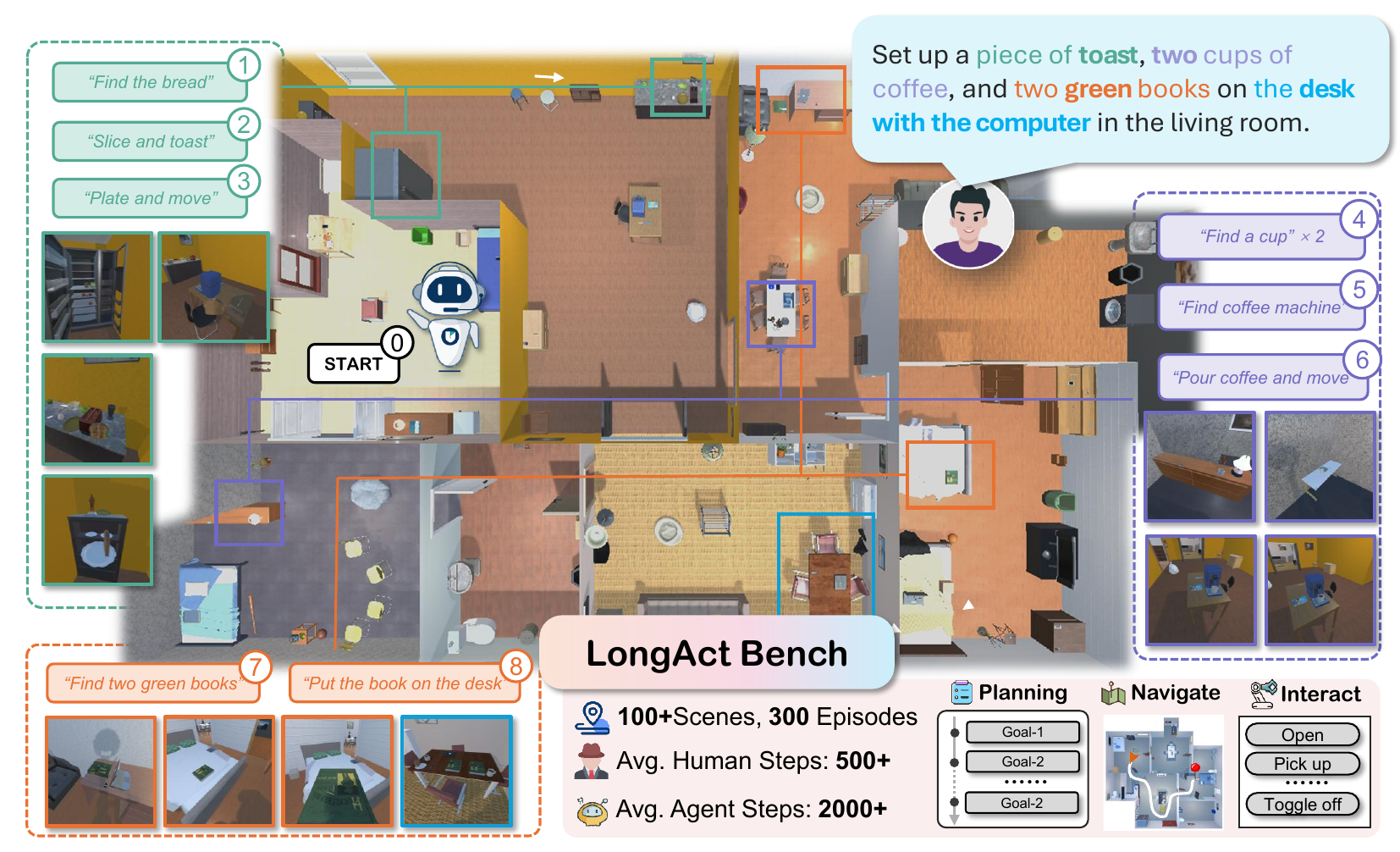}
    \captionof{figure}{\textbf{Overview of LongAct Bench.} LongAct evaluates agents on long-horizon household tasks that span 500+ human steps and require tightly coupled navigation and manipulation across multi-room environments. The benchmark emphasizes persistent reasoning, memory, and error recovery over thousands of actions, revealing the challenges of achieving reliable long-horizon autonomy.}
    \label{fig:overview}
\end{figure*}
% % -------------------------------------------------------------------------------

\section{Introduction}
\label{sec:intro}
Developing embodied agents that can autonomously carry out long-horizon tasks has become a central goal in robotics and embodied AI, with household robotics being one of the most compelling application domains. Recent advances in vision-language models (VLMs) \cite{yang2025qwen3,achiam2023gpt,Liu2023,bai2023qwenvl,team2023gemini, wang2025internvl3} have substantially improved an agent’s ability to interpret natural language instructions and carry out short sequences of actions, such as executing local manipulations or navigating to nearby objects. Supported by specialized navigation and manipulation controllers, these systems can now accomplish single-step or short-horizon tasks (e.g., “Pick up the cup”).

However, most existing datasets and benchmarks focus on relatively short-horizon tasks. Popular evaluation suites in navigation \cite{qi2020reverie, anderson2018vision,yokoyama2024hm3d, khanna2024goat,chen2020soundspaces,batra2020objectnav,CNav2025}, interactive manipulation \cite{mu2024robotwin}, and household simulation \cite{shridhar2020alfred, shridhar2020alfworld, batra2020rearrangement, gan2022threedworld} typically focus on simple objectives that involve limited temporal dependencies and modest reasoning requirements. While these tasks have driven rapid progress, they fail to capture the long, complex, and often implicitly ordered routines that characterize real household activities. In everyday settings, humans assign robots compound goals such as organizing a desk, resetting a kitchen, or performing a sequence of cleaning operations — tasks that require the agent to coordinate multiple sub-goals, manipulate shared objects, maintain spatial and semantic memory, and continuously re-plan as the environment evolves. Executing such workflows also demands that agents improve over time: learning from past mistakes, recognizing recurring patterns, and becoming more efficient as they accumulate experience. Yet these abilities remain largely unmeasured and unsupported by existing benchmarks and agent frameworks.

To bridge this gap, we introduce \textbf{LongAct}, the first agent benchmark designed for long-horizon free-form household task execution. LongAct centers on multi-stage, context-rich tasks that require continuous navigation, object manipulation, sequential decision-making, and sustained reasoning. The benchmark is built to capture the structural and procedural complexity characteristic of household workflows, using diverse multi-room simulated environments \cite{kolve2017ai2} and long-horizon task instructions that unfold over thousands of execution steps. Crucially, LongAct emphasizes autonomy at the level of task planning and reasoning. By abstracting away embodiment-specific low-level action execution, the benchmark focuses on evaluating the agent’s cognitive abilities, including interpreting instructions, maintaining state across long temporal spans, resolving dependencies or conflicts, and adapting plans based on execution feedback. This abstraction also enables seamless integration with mature real-world control modules, facilitating sim-to-real transfer without entangling high-level reasoning with robot-specific actuation details.
LongAct evaluates the fidelity and efficiency of long-horizon execution and further introduces an Improvement Rate metric that quantifies performance gains achieved through accumulated experience.

To tackle the challenges posed by LongAct and enable robust execution of complex household tasks, we introduce \textbf{HoloMind} (\textbf{Ho}rizon-\textbf{Lo}ng \textbf{Mind}), an agent framework explicitly designed for long-horizon autonomy. HoloMind employs a hierarchical planning architecture in which a high-level planner decomposes instructions into a directed acyclic graph (DAG) \cite{digitale2022tutorial} of goals, enabling structured dependencies and flexible reordering. Each goal is further grounded by a low-level planner that yields executable sub-goals. To support persistent reasoning in partially observed spaces, we design a Multimodal Spatial Memory that stores both semantic and visual features, enabling fine-grained multimodal queries and continual reconstruction of world state as the agent moves. Complementing this, an Episodic Memory compresses past experiences and distills reusable knowledge, allowing the agent to leverage lessons from past mistakes to accelerate future decisions. Finally, a Critic oversees the entire system, providing cross-module supervision and targeted corrections that 
enable self-recovery, experience reuse, and stable long-term execution.

To assess HoloMind’s effectiveness, we evaluate it on LongAct Bench with a wide range of state-of-the-art VLMs, including GPT-5 and multiple scales of Qwen3-VL (2B–32B) \cite{bai2023qwenvl}. The results highlight both the challenge posed by LongAct and the advantages of our agent design. 
First, naive VLM deployment exhibits strong scale dependence (0.74\% vs. 6.1\%, an eight-fold gap). Under HoloMind, performance rises to 25\% and 51\%, while the relative gap shrinks to roughly two-fold. This indicates that agent structure—not parameter scale alone—plays a decisive role in long-horizon execution.
Second, the Critic module also proves crucial for stable execution: removing it reduces accuracy by roughly 40\% and leads to a deterioration of up to 90\% in manipulation efficiency and Improvement Rate, demonstrating the necessity of reflective supervision during long sequences. Finally, even top-performing GPT-5 \cite{gpt5} attains only 59\% accuracy compared with 93\% for human performance, and its full-task success rate remains at 16\%. These findings confirm the difficulty of LongAct Bench and highlight the substantial room that remains for advancing long-horizon embodied intelligence. 

Our contributions are three-fold: 
(1) We propose LongAct, the first agent benchmark dedicated to long-horizon household task execution, featuring free-form natural language instructions, complex perceptual grounding, thousand-step trajectories, and an Improvement Rate metric that captures experience-driven improvement in long-horizon task execution.
(2) We introduce HoloMind, a unified long-horizon agent equipped with DAG-based hierarchical planning, persistent multimodal memory, experience accumulation, and reflective supervision to support stable autonomous execution over tasks requiring thousands of steps.
(3) Through large-scale experiments on diverse VLMs, we demonstrate that structured agent architectures can fundamentally reshape the difficulty landscape of long-horizon embodied tasks, significantly improving absolute performance while reducing rigid dependence on model scale.

\section{Related Work}
\label{sec:rel}

\textbf{Embodied AI Benchmarks.} Owing to the development of simulators for embodied agents in recent years \cite{kolve2017ai2, savva2019habitat, ramakrishnan2021habitat}, numerous embodied AI benchmarks have been proposed \cite{shridhar2020alfred, shridhar2020alfworld, batra2020rearrangement, gan2022threedworld, puig2020watch, choi2024lota, CNav2025, goatbench}. 
Although these benchmarks encompass diverse task types and environments, most are limited to relatively short task horizons. %(\cref{tab:vln_benchmarks})
The work most closely related to ours is Behavior-1K~\cite{behavior1k}, which defines 1,000 common household needs and provides detailed long-horizon task annotations for 50 categories. It represents an important step toward long-horizon household task modeling. However, evaluation on this benchmark often requires substantial domain-specific knowledge injection, and task objectives are specified through predefined goal categories rather than free-form natural language requests, limiting its suitability for benchmarking embodied agents expected to operate in real household settings, where users may issue arbitrary, free-form instructions at any time.
To evaluate agents’ planning and memory capabilities under free-form, long-horizon settings, we introduce LongAct Bench, in which each task requires over 500 steps for humans to complete, and exceeds 2,000 steps for agents on average. To the best of our knowledge, LongAct Bench is the first interactive long-horizon task planning benchmark in embodied settings that evaluates agents under free-form, instruction-driven household scenarios. A detailed comparison with existing benchmarks is provided in the appendix.

\textbf{Task Planning with LLMs.} Recent advances in Large Language Models (LLMs) \cite{dubey2024llama,bai2023qwen} enable robots to interpret natural language instructions and generate action plans through LLM-based reasoning \cite{chen2023llm, rana2023sayplan, singh2022progprompt}. While effective for simple scenarios, they often struggle with complex environments and long-horizon tasks. 
Existing methods typically rely on task-specific fine-tuning \cite{min2021film, zhao2024epo, chen2025robogpt}, which limits generalization, or few-shot in-context learning \cite{song2023llm, kim2023context, lin2025flowplan}, which can be unstable for long-term execution. Moreover, LLMs or VLMs are often used only as auxiliary modules for task decomposition or reflection, rather than forming a unified closed-loop system that integrates perception, planning, memory, and self-correction.
To address these limitations, we propose \textbf{HoloMind}, an agentic embodied framework that enables closed-loop interaction among perception, hierarchical planning, memory, and reflective supervision, supporting stable long-horizon execution and strong generalization, and can provide supervision for lightweight downstream models.

%% --------------------------------------------------------------------------
\begin{figure*}[t]
    \centering
    \includegraphics[width=\linewidth]{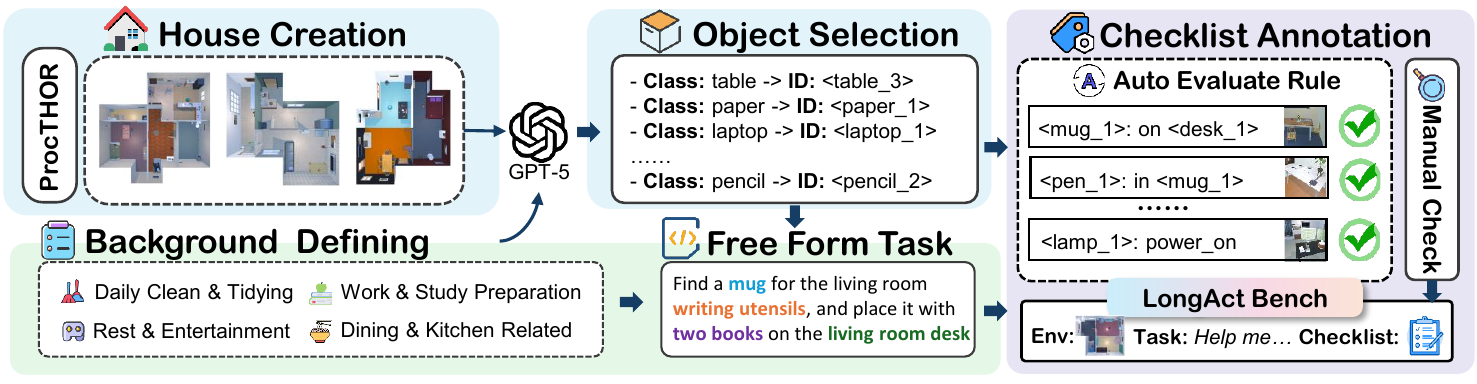}
    \vspace{-0.5em}
    \caption{
    \textbf{Overview of the LongAct benchmark construction pipeline.} Each episode includes a multi-room house environment, a multi-goal long-horizon task, and a final-state checklist for evaluation.
    }
    \vspace{-5mm}
    \label{fig:data_pipeline}
\end{figure*}
%% --------------------------------------------------------------------------

\section{The LongAct Benchmark}
\label{sec:dataset}

Considering that existing task planning datasets \cite{shridhar2020alfred, shridhar2020alfworld, puig2020watch, behavior1k} are limited by short task horizons or fixed task categories, we propose the \textbf{LongAct} benchmark to evaluate planning-level autonomy in long-horizon household task execution under free-form task descriptions. As illustrated in \cref{fig:data_pipeline}, the dataset is constructed through a collaborative pipeline involving humans and MLLMs, followed by a final manual verification stage to ensure quality.

\subsection{Long-Horizon Free-Form Task Design}
\textbf{Environment and Task Definition.}
LongAct Bench is built on diverse multi-room indoor environments generated with ProcTHOR \cite{procthor} on AI2-THOR \cite{ai2thor}. We first create $100+$ house layouts that offer varied geometry, spatial connectivity, and object distributions, enabling long-range navigation and multi-stage interactions. The benchmark comprises 300 long-horizon episodes distributed across four realistic scenarios—Daily Cleaning \& Tidying, Work \& Study Preparation, Rest \& Entertainment, and Dining \& Kitchen Related.

Each episode instruction defines a compound household routine composed of an average of 9 goals. We provide two instruction splits: a detailed split for evaluating fundamental long-horizon planning capability, and a concise split resembling natural human requests that requires intent inference through embodied interaction.
Instructions are expressed as free-form household task descriptions (e.g., “Study setup by placing the laptop and a mug of writing utensils on the bedroom desk, bringing a cup of coffee, and turning on the desk lamp.”) that specify high-level goals without prescribing explicit execution procedures. 
This design emphasizes the agent’s ability to understand task intent, generalize across diverse task compositions, and adapt its behavior through embodied exploration. In many cases, key details cannot be inferred from the instruction alone and must be grounded through interaction with the environment, requiring the agent to integrate perception, memory, and planning during execution. In addition, instructions may refer to objects at different levels of specificity. An entity may be\textbf{ identified by its category name}, or further \textbf{specified by attributes} such as color or spatial relations, requiring the agent to disambiguate among multiple candidates in the environment. 
 While human executors typically require more than 500 steps to complete a task, VLM-based agents often exceed 2,000 steps, underscoring the long-horizon challenge posed by LongAct.

\textbf{Observations and Action Space.}
Since depth estimation and semantic segmentation are well-studied perceptual subproblems with mature solutions, we provide RGB-D observations and semantic segmentation signals to decouple low-level perception from high-level reasoning and long-horizon planning, following common practice in prior embodied benchmarks \cite{goatbench}. 
For action modeling, we adopt action space defined in ALFRED \cite{shridhar2020alfred}, where the agent can execute 8 navigation actions: {\fontfamily{qcr}\selectfont MoveAhead, MoveBack, MoveLeft, MoveRight, RotateRight, RotateLeft, LookUp, LookDown}, and 7 manipulation actions: {\fontfamily{qcr}\selectfont Pick/Place, Open/Close, ToggleOn/ToggleOff} and {\fontfamily{qcr}\selectfont Slice}. Rotation and movement are set to 30\textdegree\ and 0.05m for finer control. The agent terminates with {\fontfamily{qcr}\selectfont Stop}, and each episode is capped at 16,000 steps.

\textbf{Human–MLLM Collaborative Data Construction.}
We employ a semi-automatic pipeline to construct coherent, diverse long-horizon episodes. For each selected environment and scenario, GPT-5 proposes a list of relevant objects and their desired states. Human annotators then design multi-goal workflows grounded in the proposed object set and create a detailed checklist specifying the target configuration of all relevant objects at task completion. A second team of annotators verifies and refines the tasks to ensure consistency and quality. This pipeline enables scalable creation of long-horizon, procedurally grounded household routines while maintaining realism required for robust evaluation.

\subsection{Automated Evaluation Metrics}
LongAct Bench leverages checklists to facilitate automated evaluation of agents across three key dimensions: task completion, efficiency, and self-improvement.
The checklist monitors several key object states in the environment, including {\fontfamily{qcr}\selectfont position} states, door {\fontfamily{qcr}\selectfont open/close} states, switch {\fontfamily{qcr}\selectfont on/off} states, {\fontfamily{qcr}\selectfont slicing} states, {\fontfamily{qcr}\selectfont cooking} states, and {\fontfamily{qcr}\selectfont liquid-filled} states.

\textbf{Task Completion.} A trajectory is considered successful when the states and positions of all objects in the checklist meet the requirements, from which the full-task \textit{Success Rate} (\textbf{SR}) is computed. We also introduce \textit{Goal-Condition Success} (\textbf{GC}) \cite{shridhar2020alfred} to evaluate task completion, defined as the proportion of objects whose states and positions satisfy the checklists out of the total objects number. These two metrics reflect task completion at different levels.

\textbf{Efficiency.} We directly use the number of steps required to complete a task as the metric for efficiency. To balance the contributions of navigation and manipulation, we count one step for every 0.25 meters of movement, each 30° rotation, and each object interaction when computing this metric.

\textbf{Self-improvement.} 
The agent is expected to be capable of continuously improving its performance during long-horizon task execution. To this end, we propose a new metric, \textit{Improvement Rate} (\textbf{IR}), which captures the trend in the agent’s scoring efficiency over the course of the task. Let the score sequence $\{s_t\}_{t=0}^T$ denote the accumulated GC score over a trajectory of length $T$, $\mathcal{L}(\cdot)$ denote the operator that returns the slope of the best linear least-squares fit of its inputs.
For each segmentation number $k \in \{2, \dots, n\}$, we evenly partition the score sequence into $k$ contiguous segments $S_{k,j}=\{s_t : t\in I_{k,j}\}$.
At each segmentation level, we compute the local improvement rates $a_k$:
\begin{equation}
\alpha_{k,j} = \mathcal{L}(S_{k,j}), \qquad
a_k = \mathcal{L}({\alpha_{k,1},\dots,\alpha_{k,k}}).
\end{equation}
Then IR is defined as \cref{eq:ir}. Larger values indicate stronger acceleration in improvement
\begin{equation}
I = \frac{1}{n-1} \sum_{k=2}^{n} a_k.
\label{eq:ir}
\end{equation}

%% -------------------------------------------------------------------------------
\begin{figure*}[t]
    \centering
    \includegraphics[width=\linewidth]{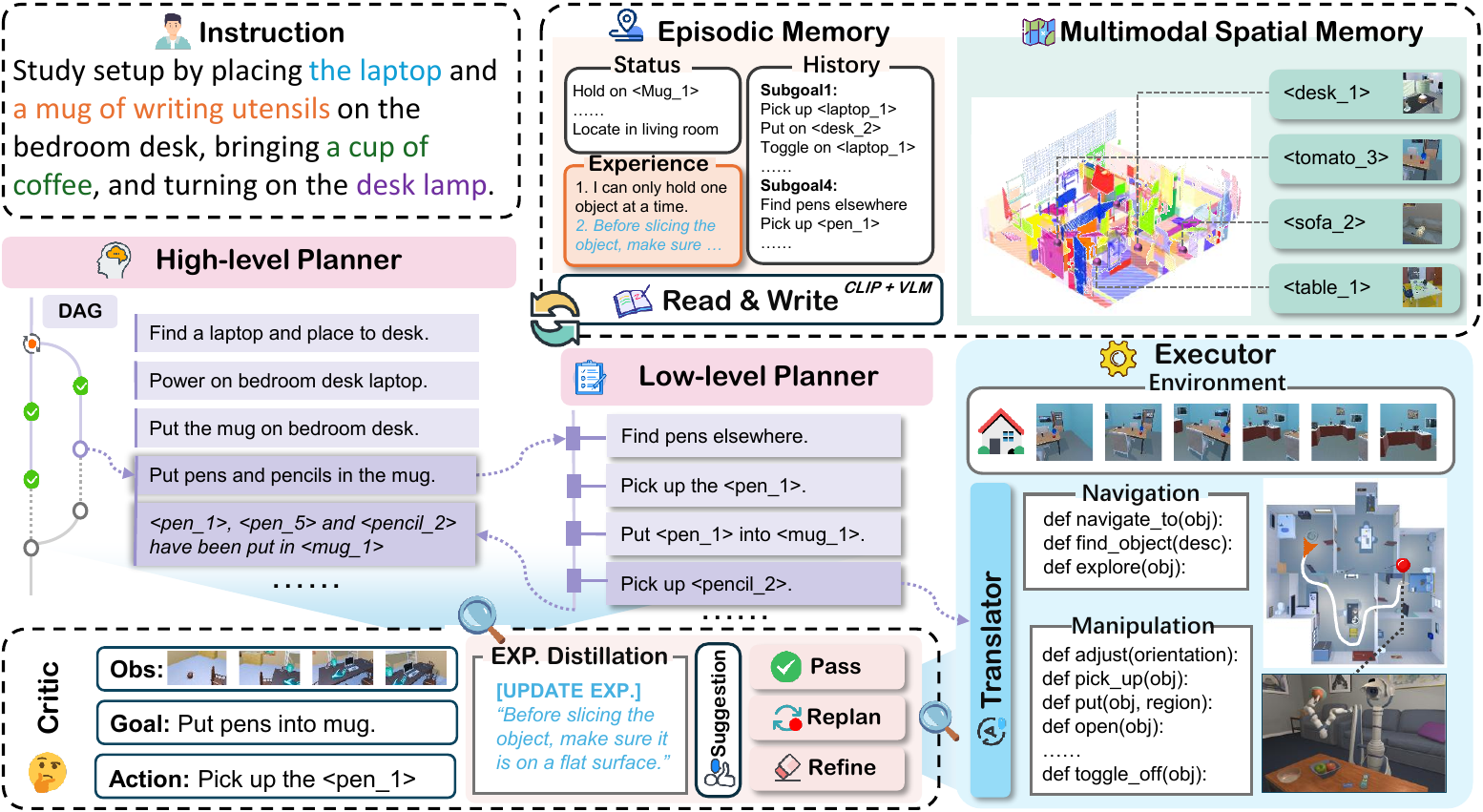}
    \caption{\textbf{Overview of the HoloMind framework.} HoloMind consists of four modules: a two-layer \textbf{Planner}, an \textbf{Executor}, a \textbf{Memory}, and a \textbf{Critic}. The \textbf{Planner} takes the task $T$ and, with support from \textbf{Memory}, incrementally decomposes it into executable natural-language instructions $I_a$. The \textbf{Executor} uses navigation and manipulation skill libraries to convert $I_a$ and observations $O_i$ into atomic simulator actions $A$. Throughout execution, the \textbf{Critic} monitors both Planner and Executor, detecting errors and providing suggestions $S$ to maintain robust closed-loop behavior.}
    \vspace{-1.0em}
    \label{fig:holo_mind}
\end{figure*}
%% -------------------------------------------------------------------------------

\section{The HoloMind Agent}

HoloMind is an embodied agent framework for zero-shot execution of complex long-horizon tasks. All core modules—including hierarchical planning, subgoal grounding, memory updating, and the reflective critic—are driven by VLM reasoning. Operating in a closed-loop manner (Fig.~\ref{fig:holo_mind}), the agent decomposes instructions into actionable subgoals, executes them while maintaining persistent memory, and continually refines its behavior through VLM-based feedback and accumulated experience.

\noindent\subsection{High-level Planner: DAG Task Decomposition}

HoloMind decomposes long-horizon tasks and enables self-optimization through a high-level planner, which operates in two key stages: DAG-based Task Decomposition, followed by Instruction Refinement through Memory.

\textbf{DAG-based Task Decomposition.} The high-level planner decomposes the task into a Directed Acyclic Graph (DAG), where nodes represent goals and edges encode dependencies. Dependent tasks are executed sequentially, while independent ones run in parallel to maximize efficiency. The planner minimizes inter-task dependencies—such as temporal, causal, object-level, or resource conflicts—and prioritizes spatially proximate goals to reduce navigation overhead, shorten execution time, and enhance both efficiency and robustness.

\textbf{Sub-Instruction Refinement through Memory.} After task decomposition, the High-level Planner refines the sub-instructions using the Memory \( M_{ep} \). Given the goal sequence \( I^{obj} \), and assuming that the goals \( \{i^{obj}_1, \dots, i^{obj}_k\} \) have been completed, the planner updates the next goal \( i^{obj}_{k+1} \) before passing it to the Low-level Planner. This update process leverages historical execution traces and contextual information from the Memory to ensure that no residual dependencies remain. Additionally, the instruction is enriched with specific details to make it more actionable. The resulting goal \( i^{r}_{k+1} \) is self-contained, object-centered, and free from unresolved dependencies, ensuring reliable execution by the Low-level Planner. This continuous feedback loop allows the High-level Planner to adapt and refine high-level instructions based on real-time execution outcomes.

\subsection{Low-level Planner: Subgoal Grounding}

The Low-level Planner specifies a detailed execution plan for each refined goal provided by the High-level Planner. For the current $k$-th goal $i^{r}_k$, the Low-level Planner identifies all objects whose spatial locations must be determined, denoted as $\{OBJ_i\}_{i=1}^l$. It first attempts to retrieve each object through the Multimodal Spatial Memory. For objects that cannot be found in memory, the Planner generates sub-goals that instruct the Executor either to directly search for the missing object or to explore likely receptacles as inferred from context. Once sufficient information has been gathered, the Planner constructs a sub-goal list for the Executor to follow in sequence. Throughout this process, the Planner must also incorporate feedback from the Critic, adjusting sub-goals when necessary to ensure stable and successful task execution.

\subsection{Executor: Robust Action with Self-Recovery}
The \textbf{Executor} serves as the interface between model decisions and the simulator. It is equipped with a \textbf{Translator} built on a VLM and a skill library composed of two parts: navigation and manipulation. The skill library provides a set of callable functions, each mapped to atomic operations in the simulator. When receiving a sub-goal produced by the Planner, the Executor uses the Translator to select and compose appropriate functions from the skill library, thereby controlling the simulator to interact with the environment. 
Under Critic oversight, the Executor performs lightweight self-recovery: when an action fails, the Critic issues brief corrective guidance or requests a fallback to the low-level Planner. This enables fast error correction without disrupting the overall hierarchy.

\subsection{Multimodal Spatial Memory: Scene Map}
Robust long-horizon execution requires an agent to maintain consistent, interpretable, and queryable world models under partial observability. To support such capabilities, HoloMind employs a Multimodal Spatial Memory that integrates geometric structure, object-level semantics, and multi-view visual evidence into a unified representation. This memory enables grounding, retrieval, disambiguation, and continual scene reconstruction, all of which are essential for VLM-driven planners in open-vocabulary settings.

The memory consists of a 3D semantic map \( M \in \mathbb{R}^3 \) \cite{vlmap, huang23avlmaps} and an object-centric database \( \{obj_i\}_{i=1}^{n} \). It stores multimodal information for each object, including multi-view RGB images \( \{I_{i,j}\}_{j=1}^{v} \), occupied voxels \( \{V_i\} \subset M \), segmentation masks \( I_i^{seg} \), and textual labels \( T_i \). To support fast retrieval, we build a dual index: a spatial index that maps objects to their voxel regions, and a semantic index constructed via CLIP embeddings \cite{radford2021learning}, which facilitates subsequent object merging and retrieval.
\begin{equation}
f_i = \text{CLIP}(I_i^{seg}) + \text{CLIP}(T_i),
\end{equation}
During exploration, when a new object \( obj_n \) is observed, the agent queries the semantic map to search for the most similar existing object using \( \{V_n\} \) and \( T_n \). If a match is found, their information is merged, and the semantic index is updated as 
$f_i = \alpha f_i + (1 - \alpha) f_n,$ where \( \alpha \) is determined by the ratio of the observed surface areas between the two views. If no match is found, a new object entry is created. When the Planner queries the Memory with a text description \( T_q \), retrieval is performed by ranking objects based on semantic similarity \( \mathrm{sim}(f_{T_q}, f_i) \) for \( i = 1,\dots,n \), and the top-\( k \) relevant candidate objects are selected. For each candidate, a VLM is applied to \( \{I_{i,j}\}_{j=1}^{v} \) to verify whether the object satisfies the description \( T_q \) , resolving ambiguities that require spatial reasoning.

\subsection{Episodic Memory: Experience Abstraction}
To enable context-aware reasoning over thousands of actions, HoloMind incorporates an Episodic Memory that captures the agent’s evolving interaction history and operational experience. Episodic memory maintains dynamic status, past decisions, and distilled experience that directly inform high-level planning. The Status component maintains real-time information, including the object currently held by the agent and its current spatial position in the environment. The History component records past actions and experiences as a sequence of completed subgoals. This historical trace helps the system avoid redundant behaviors and provides valuable context for subsequent planning.

Furthermore, the Experience component encodes learned reusable rules, constraints, and learned priors, such as the limitation that only one object can be held at a time  or that objects must be placed on a flat surface prior to manipulation. These components of Episodic Memory work in tandem to refine the system's ability to plan by referencing both current and historical states, thereby ensuring robust, context-aware decision-making in complex environments.

\subsection{Critic: Cross-Module Reflective Oversight}
The Critic provides continuous cross-module oversight throughout execution, offering feedback, corrective signals, and improvement suggestions to the High-level Planner, Low-level Planner and the Executor. Rather than directly modifying internal modules, the Critic makes supervisory decisions that steer planners and executors toward stable long-horizon behavior.

\textbf{Closed-Loop Supervision.}
The Critic continuously evaluates execution by integrating real-time observations with latent states from the planners and the executor. In addition to high-level supervisory commands, it also generates textual diagnostic feedback that explains the suspected issue or suggests alternative actions. At each step, the Critic issues one of three directives:
(i) \textit{Pass}, which means to proceed as planned;
(ii) \textit{Refine}, which triggers local adjustments within the current module, such as changing viewpoint or action parameters;
(iii) \textit{Replan}, which initiates higher-level replanning by backtracking to upstream modules.
For example, upon receiving the instruction “Pick up the apple on the table,” if the apple is not visible, the Critic first instructs the Executor to adjust its orientation for better perception. If the object remains undetected, the Critic escalates the failure to the high-level Planner for task-level reassessment.

\textbf{Experience Distillation.}
To avoid repeated mistakes, Critic maintains an experience pool that stores past failure cases and their successful resolutions. Upon encountering a familiar situation, it retrieves and applies the corresponding fix directly, bypassing redundant trial-and-error. Beyond providing immediate fixes, the Critic also distills validated resolutions back into the agent’s Episodic Memory, updating the global repository of reusable knowledge maintained across episodes, thereby enabling the agent to accumulate long-term competence.

\section{Experiments}
\label{sec:exp}

\begin{table*}[t]
\caption{\textbf{Performance comparison on LongAct Bench detailed split. } 
GC denotes goal completion. PP: \textit{Pick/Place}; TO: \textit{ToggleOn/Off}; 
OC: \textit{Open/Close}; Sl.: \textit{Slice}. 
$\dagger$ indicates pure VLM baselines without agentic architecture. }
\label{tab:scaling}
\centering
\begingroup
\setlength{\tabcolsep}{3pt}
\small
\begin{tabular}{lccccccccc}
\toprule
& \multicolumn{5}{c}{GC $\uparrow$ (\%)} & SR $\uparrow$ (\%) & \multicolumn{2}{c}{Steps $\downarrow$} & IR $\uparrow$ \\
\cmidrule(lr){2-6} \cmidrule(lr){8-9}
Model & Avg. & PP & TO & OC & Sl. &  & Nav & Manip &  \\
\midrule
\multicolumn{10}{c}{\textbf{Pure VLM}} \\
\midrule
Qwen3-VL-8B$^\dagger$  & 0.74  & 0.15  & 6.93  & 0.00   & 0.00  & 0.00  & 2598   & 0.26  & -0.08   \\
Qwen3-VL-32B$^\dagger$ & 6.14 & 3.54 & 18.8 & 7.32 & 0.00 & 0.00 & 2981 & 3.10 & -0.32 \\
\midrule
\multicolumn{10}{c}{\textbf{Agentic (HoloMind)}} \\
\midrule
GPT-5            & \textbf{59.0} & \textbf{66.7} & \textbf{82.9} & \textbf{75.2} & \textbf{60.0} & \textbf{16.0} & 1982 & 25.3 & \textbf{1.70} \\
GPT-5-mini       & 38.4 & 40.5 & 58.9 & 66.7 & 25.0 & 9.00 & 2304 & 21.5 & 0.81 \\
Qwen3-VL-32B     & 51.2 & 52.7 & 71.2 & 73.4 & 25.0 & 15.0 & 1692 & 28.7 & 1.61 \\
Qwen3-VL-8B      & 24.5 & 27.4 & 55.0 & 48.7 & 31.3 & 3.00 & 1896 & 30.6 & 0.99 \\
Qwen3-VL-2B      & 7.16 & 7.04 & 8.77 & 37.8 & 0.00 & 0.00 & 844  & 4.81 & 0.59 \\
\bottomrule
\end{tabular}
\endgroup
\end{table*}

\subsection{Main Results}

\textbf{How Does Model Scale Interact with Agent Structure?} 
Table~\ref{tab:scaling} summarizes performance across different backbone models. Under the HoloMind framework, model scale remains an important factor: larger backbones consistently achieve higher GC, SR, and IR. For instance, GPT-5 reaches 59.0\% GC and 16.0\% SR, outperforming smaller Qwen3-VL variants. These results indicate that larger backbone capacity directly contributes to more reliable long-horizon task execution, and confirm that LongAct imposes temporal and structural complexity to reveal clear capability gaps between different model scales. However, the contrast with pure VLM baselines reveals a more critical insight. Without agentic structure, scaling alone provides limited gains: Qwen3-VL-32B achieves only 6.14\% GC and fails entirely in SR, while Qwen3-VL-8B collapses to 0.74\%. In contrast, when integrated into HoloMind, the same backbones reach 51.2\% and 24.5\% GC respectively. Notably, the relative performance gap between 32B and 8B shrinks substantially under the agentic framework. This demonstrates that structured task decomposition and closed-loop reasoning fundamentally reshape the difficulty landscape of long-horizon execution, mitigating rigid dependence on parameter scale.

\textbf{Improvement Rate and Long-Term Adaptation.}
The Improvement Rate (IR) further highlights the distinction between naive and structured execution. Pure VLM baselines exhibit negative IR, indicating performance degradation over time as errors accumulate across long trajectories. In contrast, all HoloMind-based agents achieve positive IR, with larger models exceeding 1.6. This consistent improvement reflects the effectiveness of persistent memory and reflective supervision in stabilizing execution and enabling experience accumulation. Rather than drifting into compounding failures, agentic systems progressively refine their decisions across interactions, which is essential for sustained long-horizon autonomy.

% -----------------------------------------------------------------------------------------
\begin{figure*}[t]
    \centering
    \includegraphics[width=\linewidth]{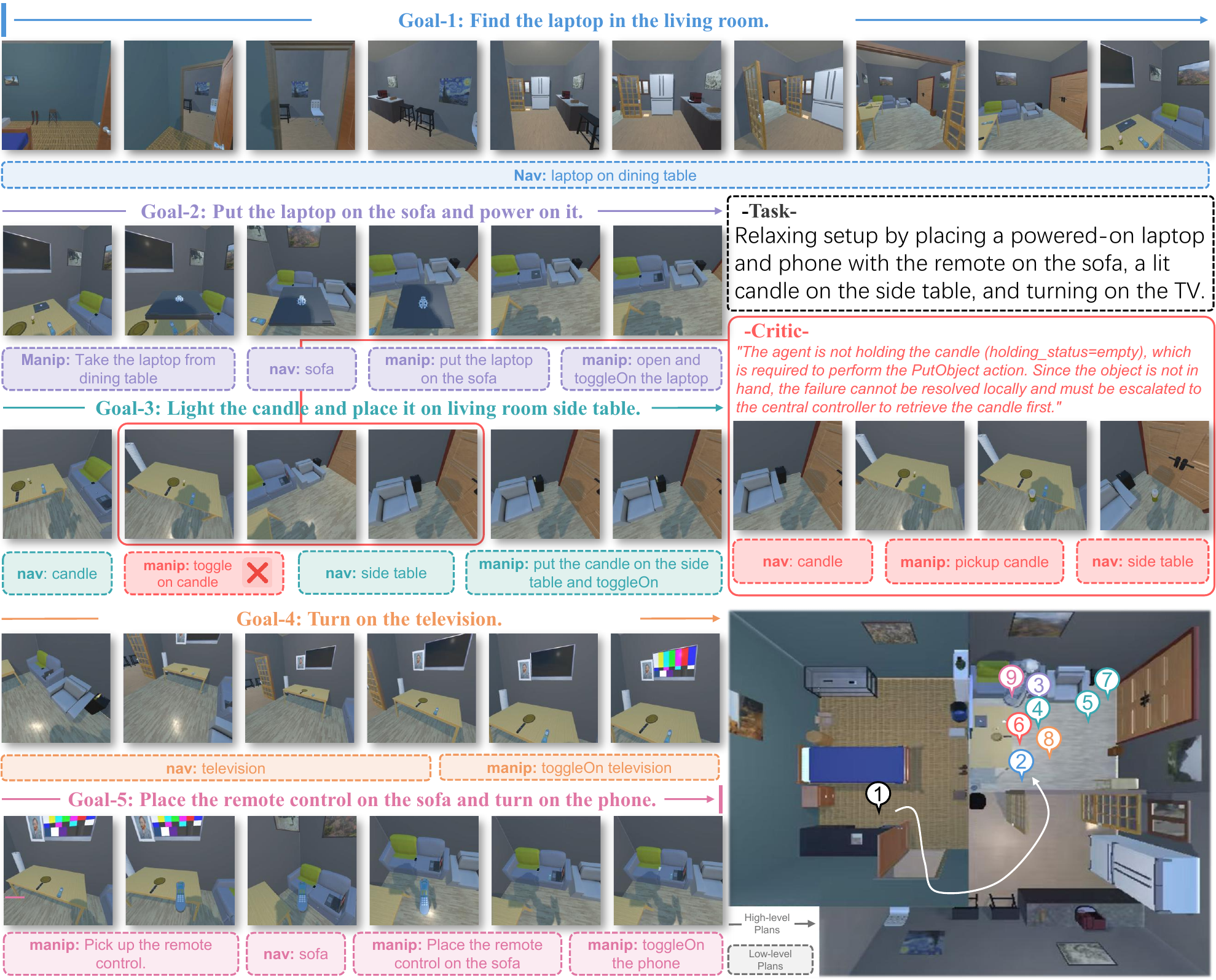}
    \vspace{-2pt}
    \caption{
    \textbf{A sample trajectory visualization of HoloMind on LongAct Bench.} The agent decomposes the task into five goals executed through interleaved navigation and manipulation steps. When encountering an issue (highlighted in red), HoloMind analyzes the cause of failure and adjusts its strategy accordingly.
    }
    \vspace{-0.7em}
    \label{fig:traj_vis}
\end{figure*}
% -----------------------------------------------------------------------------------------

\textbf{How Do Agents Act Over Time?} Fig. \ref{fig:traj_vis} shows a visualization of an agent trajectory, where the task is decomposed into five sequential goals. When encountering errors, the agent autonomously reflects on and diagnoses the issue, then adjusts its strategy accordingly. This result demonstrates the effectiveness of HoloMind and suggests that long-horizon task execution is not merely a scaling issue but a structural challenge. Additional examples are provided in Appendix.

% =========================
% Full-width table (top)
% Requires: \usepackage{booktabs}

\begin{table*}[t]
  \centering
\setlength{\tabcolsep}{3pt}
\small
    \caption{\textbf{Results on concise split.} Performance of HoloMind vs.\ the pure VLM. 
    CK and FW denote \textit{Cook} and \textit{FillWith}, respectively.}
  \label{tab:overall}
  \begin{tabular}{@{}l c c c c c c c c c c@{}}
    \toprule
    & \multicolumn{6}{c}{GC $\uparrow$ (\%)} & SR $\uparrow$ (\%) & \multicolumn{2}{c}{Steps $\downarrow$} & IR $\uparrow$ \\
    \cmidrule(lr){2-7} \cmidrule(lr){9-10}
    Model & Avg. & PP  & TO  & Sl.  & CK  & FW  & & Nav & Manip &  \\
    \midrule
    Qwen3-VL-32B & 1.84 & 0.50 & 24.0  & 0.00 & 0.00 & 0.00 & 0.00 & 1998 & 356 & -1.12 \\
    +HoloMind    & 16.2 & 10.2 & 41.3 & 8.68  & 1.39 & 2.27 & 1.00 & 6562 & 18.6 & 1.19 \\
    \bottomrule
  \end{tabular}
  \vspace{-2mm}
\end{table*}

% =========================
% Half-width table + half-width figure
% =========================
\begin{figure}[t]
  \centering

  % -------- Left (Table) --------
  \begin{minipage}[t]{0.48\linewidth}
    \vspace{0pt}
    \centering
    \small

    \captionof{table}{\textbf{Error composition analysis.} Error categories are labeled by an LLM for each failure instance.}
    \label{tab:error_comp}

    \begin{tabular}{@{}l r@{}}
      \toprule
      Category & Portion (\%) \\
      \midrule
      \textit{Task Planning} & 27.7 \\
      \midrule
      \textit{Perception/Memory} & 26.0 \\
      \quad Unrecalled & 10.7 \\
      \quad Partial    & 1.7 \\
      \quad Mismatch   & 9.7 \\
      \quad Absent     & 4.0 \\
      \midrule
      \textit{Execution} & 46.3 \\
      \bottomrule
    \end{tabular}
  \end{minipage}
  \hfill
  % -------- Right (Figure) --------
  \begin{minipage}[t]{0.48\linewidth}
    \vspace{0pt}
    \centering
    \includegraphics[width=0.58\linewidth]{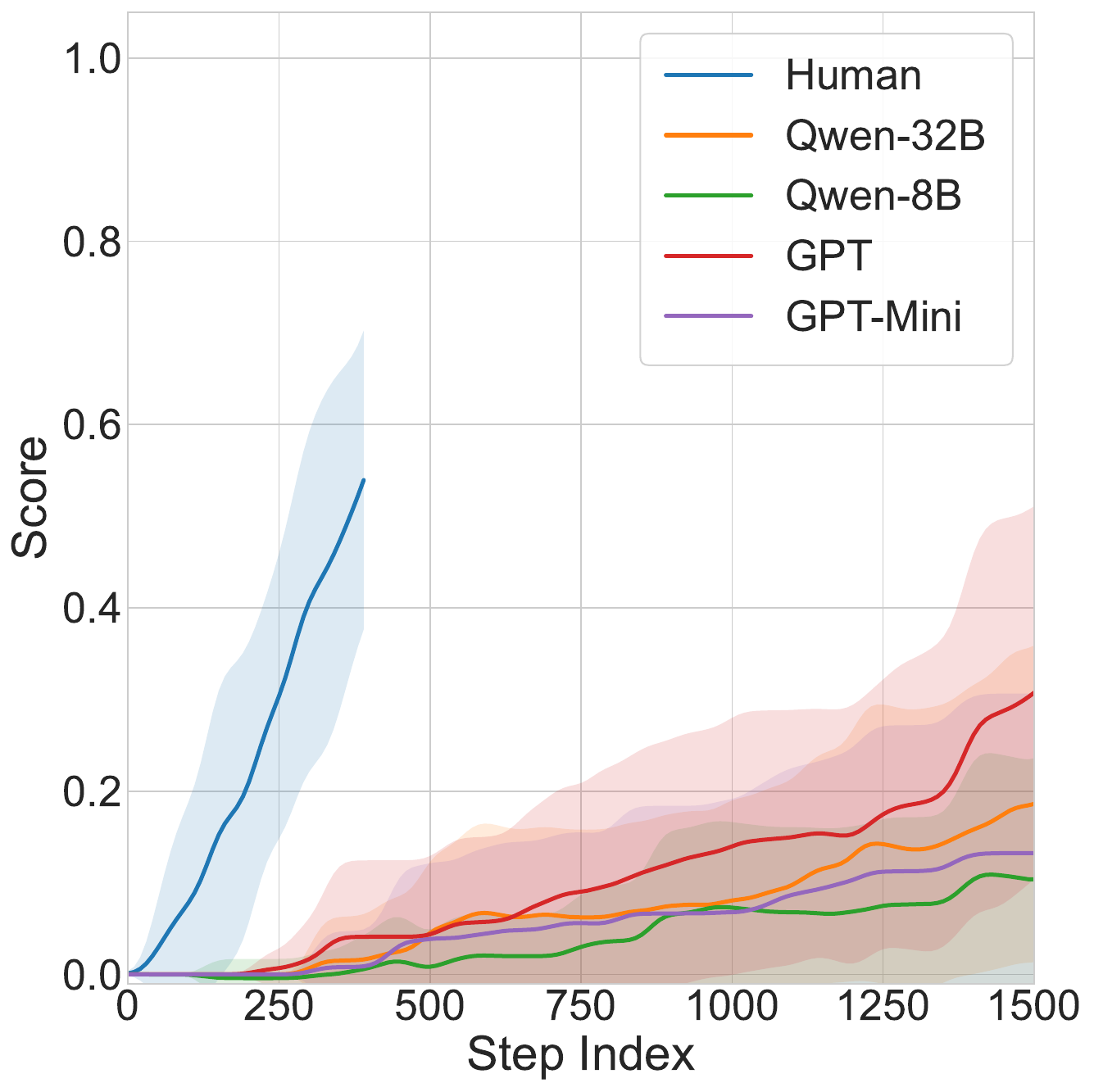}
    \vspace{-2mm}
    \caption{\textbf{Human and agent scoring trajectories Comparison.}}
    \label{fig:timeline}
  \end{minipage}

  \vspace{-2mm}
\end{figure}

\textbf{What Types of Errors Remain?}
Table~\ref{tab:error_comp} analyzes failure cases based on execution logs. Execution errors account for the largest portion (46.3\%), followed by task planning (27.7\%) and memory-related issues (26.0\%). Notably, memory failures remain frequent even with ground-truth segmentation and depth signals. Although the benchmark provides reliable low-level perception, agents must still correctly ground entities through attributes and spatial relations. For example, distinguishing the target object may require reasoning over multiple candidates that share the same category label but differ in location or appearance. These results suggest that reliable low-level perception alone does not eliminate failures in long-horizon tasks, where correct grounding and reasoning over object relations remain critical.

\textbf{Challenges of Concise Instructions.}
Performance on the Concise Split drops substantially compared to the Detailed Split, highlighting the additional difficulty of interpreting underspecified instructions. In this setting, agents must infer implicit sub-goals and dynamically ground task intent through interaction with the environment. However, under the current architecture, planning adjustments are largely reactive: agents revise plans only after encountering failures, rather than actively gathering information to guide decision-making. The absence of proactive exploration and information-seeking planning leads to cascading errors in long trajectories. This result further demonstrates the difficulty of LongAct Bench and underscores the importance of integrating active perception and online planning mechanisms for real-world embodied agents.

\subsection{Ablation Study}

\begin{table}[t]
  \centering
  \begin{minipage}[t]{0.48\linewidth}
    \centering
    \small
    \caption{\textbf{Ablation study on core components.} Starting from Qwen3-VL-32B for all modules, each variant disables exactly one module while keeping the rest of the system intact. \textit{Human-tester} denotes a human controlling the robot in the simulator via keyboard and mouse with revisitable task instructions.}
    \label{tab:module}
    \begin{tabular}{@{}lcccc@{}}
      \toprule
      Method & GC $\uparrow$ & SR $\uparrow$ & Steps $\downarrow$ & IR $\uparrow$ \\
      \midrule
      Human-tester & 92.9 & 66.7 & 376/56.5 & 1.78\\
      \midrule
      Qwen3-VL-32B & 51.2 & 15.0 & 1692/28.7 & 1.64 \\
      - HighPlanner & 32.6 & 12.2 & 1435/24.5 & 0.60 \\
      - EpisMem & 51.3 & 17.7 & 2221/36.3 & 1.11 \\
      - Critic & 29.2 & 6.54 & 2018/222 & 0.16 \\
      \bottomrule
    \end{tabular}
  \end{minipage}
  \hfill
  \begin{minipage}[t]{0.48\linewidth}
    \centering
    \small
    \caption{\textbf{Ablation study by enhancing module capacity.} Starting from Qwen3-VL-8B for all modules, each row shows the effect of enhancing \emph{only one} module with VLM (32B) while keeping all other modules unchanged. $\uparrow$ indicates enhanced module.}
    \label{tab:results}
    \begin{tabular}{@{}lcccc@{}}
      \toprule
      Method & GC $\uparrow$ & SR $\uparrow$ & Steps $\downarrow$ & IR $\uparrow$ \\
      \midrule
      Qwen3-VL-32B & 51.2 & 15.0 & 1692/28.7 & 1.64 \\
      \midrule
      Qwen3-VL-8B & 24.5 & 2.80 & 1896/30.6 & 0.99 \\
      + HighPlanner$\uparrow$ & 26.3 & 2.80 & 1752/32.1 & 1.32 \\
      + LowPlanner$\uparrow$ & 42.8 & 7.47 & 1783/33.3 & 1.41 \\
      + Critic$\uparrow$ & 28.5 & 6.54 & 2049/25.5 & 1.10 \\
      + Executor$\uparrow$ & 27.1 & 4.67 & 1879/33.0 & 1.31 \\
      \bottomrule
    \end{tabular}
  \end{minipage}
  \vspace{-2mm}
\end{table}

\textbf{How Much Does Each Module Contribute?} 
The ablation results in Table~\ref{tab:module} highlight the distinct roles of each module in HoloMind. Removing the HighPlanner reduces GC by 18.6\% and SR by 3.8\%, with IR dropping by 1.04, showing that hierarchical decomposition is critical for coherent long-horizon reasoning. 
Removing Episodic Memory has little effect on GC (+0.1\%) and slightly improves SR (+2.7\%), but significantly increases navigation (+529) and manipulation steps (+7.6), while IR drops by 0.53. This indicates that episodic memory mainly improves execution efficiency by reducing redundant exploration rather than affecting decision accuracy.
Removing the Critic causes the most severe degradation: GC drops by 22.0\%, SR by 8.46\%, manipulation steps surge by 193.3, and IR decreases by 1.48. This confirms that reflective supervision is essential for stabilizing the Planner–Executor pipeline and preventing cascading errors in long-horizon execution.

\textbf{Which Module Needs More Capacity?}
The ablation study in Table \ref{tab:results} evaluates how upgrading individual modules from Qwen3-VL-8B to Qwen3-VL-32B affects system performance. Strengthening the LowPlanner yields the largest gains, increasing GC by 18.3\%, SR by 4.67\%, and IR by 0.42. This indicates that the LowPlanner plays a critical role in grounding high-level goals into executable subgoals and generating reliable action sequences.
Enhancing the Critic improves SR by 3.74\% along with modest gains on other metrics, highlighting its important role in maintaining robust execution. Upgrading the High-level Planner or Executor brings only small improvements across metrics, suggesting that the overall task decomposition design of HoloMind is well-balanced.

\subsection{Compared with Human Performance}

To better understand how agents accumulate progress over long trajectories, we design a scoring--timeline experiment comparing model behavior with human performance as an upper-bound reference. As shown in Fig.~\ref{fig:timeline}, the results plot score trajectories over the first 1500 steps (only trajectories with total steps $>$1500 are retained to avoid early-termination artifacts in visualization). A clear capacity-dependent separation emerges: humans accumulate scores rapidly and consistently while model agents exhibit slower and capacity-dependent progress. The four HoloMind agents powered by different VLMs exhibit similar overall trends but with varying magnitudes, providing additional evidence for the soundness of the HoloMind architecture. Stronger models demonstrate increasingly larger advantages in later stages, aligning with intuitive expectations and further supporting the validity of the IR metric.

% human

\section{Conclusion}
\label{sec:conclusion}
We presented LongAct, a benchmark that shifts embodied evaluation toward the long-horizon planning and persistent reasoning demanded by real household tasks. By isolating high-level autonomy from low-level control, LongAct exposes the limitations of current VLM-based agents and reveals the widening gap between short-horizon competence and the sustained reasoning needed for real-world tasks. Building on this perspective, we introduced HoloMind, a hierarchical and memory-centric agent designed to sustain coherent plans, manage evolving world states, and learn from extended experience. 
While HoloMind offers clear gains, substantial headroom remains, pointing to long-horizon reasoning as a central open challenge for embodied AI.

%Bibliography
\bibliographystyle{unsrt}  
\bibliography{references}  

@inproceedings{radford2021learning,
  title={Learning transferable visual models from natural language supervision},
  author={Radford, Alec and Kim, Jong Wook and Hallacy, Chris and Ramesh, Aditya and Goh, Gabriel and Agarwal, Sandhini and Sastry, Girish and Askell, Amanda and Mishkin, Pamela and Clark, Jack and others},
  booktitle={International conference on machine learning},
  pages={8748--8763},
  year={2021},
  organization={PmLR}
}

@inproceedings{huang23avlmaps,
              title={Audio Visual Language Maps for Robot Navigation},
              author={Chenguang Huang and Oier Mees and Andy Zeng and Wolfram Burgard},
              booktitle={Proceedings of the International Symposium on Experimental Robotics (ISER)},
              year={2023},
              address = {Chiang Mai, Thailand}
          }

@inproceedings{mu2024robotwin,
  title={Robotwin: Dual-arm robot benchmark with generative digital twins (early version)},
  author={Mu, Yao and Chen, Tianxing and Peng, Shijia and Chen, Zanxin and Gao, Zeyu and Zou, Yude and Lin, Lunkai and Xie, Zhiqiang and Luo, Ping},
  booktitle={European Conference on Computer Vision},
  pages={264--273},
  year={2024},
  organization={Springer}
}

@article{bai2023qwen,
  title={Qwen technical report},
  author={Bai, Jinze and Bai, Shuai and Chu, Yunfei and Cui, Zeyu and Dang, Kai and Deng, Xiaodong and Fan, Yang and Ge, Wenbin and Han, Yu and Huang, Fei and others},
  journal={arXiv preprint arXiv:2309.16609},
  year={2023}
}

@article{achiam2023gpt,
  title={Gpt-4 technical report},
  author={Achiam, Josh and Adler, Steven and Agarwal, Sandhini and Ahmad, Lama and Akkaya, Ilge and Aleman, Florencia Leoni and Almeida, Diogo and Altenschmidt, Janko and Altman, Sam and Anadkat, Shyamal and others},
  journal={arXiv preprint arXiv:2303.08774},
  year={2023}
}

@inproceedings{chen2020soundspaces,
  title={Soundspaces: Audio-visual navigation in 3d environments},
  author={Chen, Changan and Jain, Unnat and Schissler, Carl and Gari, Sebastia Vicenc Amengual and Al-Halah, Ziad and Ithapu, Vamsi Krishna and Robinson, Philip and Grauman, Kristen},
  booktitle={European conference on computer vision},
  pages={17--36},
  year={2020},
  organization={Springer}
}

@article{batra2020objectnav,
  title={Objectnav revisited: On evaluation of embodied agents navigating to objects},
  author={Batra, Dhruv and Gokaslan, Aaron and Kembhavi, Aniruddha and Maksymets, Oleksandr and Mottaghi, Roozbeh and Savva, Manolis and Toshev, Alexander and Wijmans, Erik},
  journal={arXiv preprint arXiv:2006.13171},
  year={2020}
}

@article{digitale2022tutorial,
  title={Tutorial on directed acyclic graphs},
  author={Digitale, Jean C and Martin, Jeffrey N and Glymour, Medellena Maria},
  journal={Journal of Clinical Epidemiology},
  volume={142},
  pages={264--267},
  year={2022},
  publisher={Elsevier}
}

@inproceedings{khanna2024goat,
  title={Goat-bench: A benchmark for multi-modal lifelong navigation},
  author={Khanna, Mukul and Ramrakhya, Ram and Chhablani, Gunjan and Yenamandra, Sriram and Gervet, Theophile and Chang, Matthew and Kira, Zsolt and Chaplot, Devendra Singh and Batra, Dhruv and Mottaghi, Roozbeh},
  booktitle={Proceedings of the IEEE/CVF Conference on Computer Vision and Pattern Recognition},
  pages={16373--16383},
  year={2024}
}

@inproceedings{yokoyama2024hm3d,
  title={Hm3d-ovon: A dataset and benchmark for open-vocabulary object goal navigation},
  author={Yokoyama, Naoki and Ramrakhya, Ram and Das, Abhishek and Batra, Dhruv and Ha, Sehoon},
  booktitle={2024 IEEE/RSJ International Conference on Intelligent Robots and Systems (IROS)},
  pages={5543--5550},
  year={2024},
  organization={IEEE}
}

@article{Liu2023,
  author    = {H. Liu and C. Li and Q. Wu and Y. J. Lee},
  title     = {Visual Instruction Tuning},
  journal   = {Advances in Neural Information Processing Systems},
  volume    = {36},
  pages     = {1--19},
  year      = {2023},
}

@article{wang2025internvl3,
  title={Internvl3. 5: Advancing open-source multimodal models in versatility, reasoning, and efficiency},
  author={Wang, Weiyun and Gao, Zhangwei and Gu, Lixin and Pu, Hengjun and Cui, Long and Wei, Xingguang and Liu, Zhaoyang and Jing, Linglin and Ye, Shenglong and Shao, Jie and others},
  journal={arXiv preprint arXiv:2508.18265},
  year={2025}
}

@article{bai2023qwenvl,
  title={Qwen-vl: A frontier large vision-language model with versatile abilities},
  author={Bai, Jinze and Bai, Shuai and Yang, Shusheng and Wang, Shijie and Tan, Sinan and Wang, Peng and Lin, Junyang and Zhou, Chang and Zhou, Jingren},
  journal={arXiv preprint arXiv:2308.12966},
  year={2023}
}

@article{dubey2024llama,
  title={The llama 3 herd of models},
  author={Dubey, Abhimanyu and Jauhri, Abhinav and Pandey, Abhinav and Kadian, Abhishek and Al-Dahle, Ahmad and Letman, Aiesha and Mathur, Akhil and Schelten, Alan and Yang, Amy and Fan, Angela and others},
  journal={arXiv preprint arXiv:2407.21783},
  year={2024}
}

@article{team2023gemini,
  title={Gemini: a family of highly capable multimodal models},
  author={Team, Gemini and Anil, Rohan and Borgeaud, Sebastian and Alayrac, Jean-Baptiste and Yu, Jiahui and Soricut, Radu and Schalkwyk, Johan and Dai, Andrew M and Hauth, Anja and Millican, Katie and others},
  journal={arXiv preprint arXiv:2312.11805},
  year={2023}
}

@article{yang2025qwen3,
  title={Qwen3 technical report},
  author={Yang, An and Li, Anfeng and Yang, Baosong and Zhang, Beichen and Hui, Binyuan and Zheng, Bo and Yu, Bowen and Gao, Chang and Huang, Chengen and Lv, Chenxu and others},
  journal={arXiv preprint arXiv:2505.09388},
  year={2025}
}

@article{CNav2025,
  title={C-NAV: Towards Self-Evolving Continual Object Navigation in Open World},
  author={Yu, Ming-Ming and Zhu, Fei and Liu, Wenzhuo and Yang, Yirong and Wang, Qunbo and Wu, Wenjun and Liu, Jing},
  journal={arXiv preprint arXiv:2510.20685},
  year={2025}
}

@inproceedings{savva2019habitat,
  title={Habitat: A platform for embodied ai research},
  author={Savva, Manolis and Kadian, Abhishek and Maksymets, Oleksandr and Zhao, Yili and Wijmans, Erik and Jain, Bhavana and Straub, Julian and Liu, Jia and Koltun, Vladlen and Malik, Jitendra and others},
  booktitle={Proceedings of the IEEE/CVF international conference on computer vision},
  pages={9339--9347},
  year={2019}
}

@article{ramakrishnan2021habitat,
  title={Habitat-matterport 3d dataset (hm3d): 1000 large-scale 3d environments for embodied ai},
  author={Ramakrishnan, Santhosh K and Gokaslan, Aaron and Wijmans, Erik and Maksymets, Oleksandr and Clegg, Alex and Turner, John and Undersander, Eric and Galuba, Wojciech and Westbury, Andrew and Chang, Angel X and others},
  journal={arXiv preprint arXiv:2109.08238},
  year={2021}
}

@article{kolve2017ai2,
  title={Ai2-thor: An interactive 3d environment for visual ai},
  author={Kolve, Eric and Mottaghi, Roozbeh and Han, Winson and VanderBilt, Eli and Weihs, Luca and Herrasti, Alvaro and Deitke, Matt and Ehsani, Kiana and Gordon, Daniel and Zhu, Yuke and others},
  journal={arXiv preprint arXiv:1712.05474},
  year={2017}
}

@article{puig2020watch,
  title={Watch-and-help: A challenge for social perception and human-ai collaboration},
  author={Puig, Xavier and Shu, Tianmin and Li, Shuang and Wang, Zilin and Liao, Yuan-Hong and Tenenbaum, Joshua B and Fidler, Sanja and Torralba, Antonio},
  journal={arXiv preprint arXiv:2010.09890},
  year={2020}
}

@article{choi2024lota,
  title={Lota-bench: Benchmarking language-oriented task planners for embodied agents},
  author={Choi, Jae-Woo and Yoon, Youngwoo and Ong, Hyobin and Kim, Jaehong and Jang, Minsu},
  journal={arXiv preprint arXiv:2402.08178},
  year={2024}
}

@inproceedings{shridhar2020alfred,
  title={Alfred: A benchmark for interpreting grounded instructions for everyday tasks},
  author={Shridhar, Mohit and Thomason, Jesse and Gordon, Daniel and Bisk, Yonatan and Han, Winson and Mottaghi, Roozbeh and Zettlemoyer, Luke and Fox, Dieter},
  booktitle={Proceedings of the IEEE/CVF conference on computer vision and pattern recognition},
  pages={10740--10749},
  year={2020}
}

@article{shridhar2020alfworld,
  title={Alfworld: Aligning text and embodied environments for interactive learning},
  author={Shridhar, Mohit and Yuan, Xingdi and C{\^o}t{\'e}, Marc-Alexandre and Bisk, Yonatan and Trischler, Adam and Hausknecht, Matthew},
  journal={arXiv preprint arXiv:2010.03768},
  year={2020}
}

@article{chen2023llm,
  title={Llm-state: Expandable state representation for long-horizon task planning in the open world},
  author={Chen, Siwei and Xiao, Anxing and Hsu, David},
  journal={CoRR},
  year={2023}
}

@article{rana2023sayplan,
  title={Sayplan: Grounding large language models using 3d scene graphs for scalable robot task planning},
  author={Rana, Krishan and Haviland, Jesse and Garg, Sourav and Abou-Chakra, Jad and Reid, Ian and Suenderhauf, Niko},
  journal={arXiv preprint arXiv:2307.06135},
  year={2023}
}

@article{singh2022progprompt,
  title={Progprompt: Generating situated robot task plans using large language models},
  author={Singh, Ishika and Blukis, Valts and Mousavian, Arsalan and Goyal, Ankit and Xu, Danfei and Tremblay, Jonathan and Fox, Dieter and Thomason, Jesse and Garg, Animesh},
  journal={arXiv preprint arXiv:2209.11302},
  year={2022}
}

@article{min2021film,
  title={Film: Following instructions in language with modular methods},
  author={Min, So Yeon and Chaplot, Devendra Singh and Ravikumar, Pradeep and Bisk, Yonatan and Salakhutdinov, Ruslan},
  journal={arXiv preprint arXiv:2110.07342},
  year={2021}
}

@article{zhao2024epo,
  title={Epo: Hierarchical llm agents with environment preference optimization},
  author={Zhao, Qi and Fu, Haotian and Sun, Chen and Konidaris, George},
  journal={arXiv preprint arXiv:2408.16090},
  year={2024}
}

@article{chen2025robogpt,
  title={Robogpt: an llm-based long-term decision-making embodied agent for instruction following tasks},
  author={Chen, Yaran and Cui, Wenbo and Chen, Yuanwen and Tan, Mining and Zhang, Xinyao and Liu, Jinrui and Li, Haoran and Zhao, Dongbin and Wang, He},
  journal={IEEE Transactions on Cognitive and Developmental Systems},
  year={2025},
  publisher={IEEE}
}

@inproceedings{song2023llm,
  title={Llm-planner: Few-shot grounded planning for embodied agents with large language models},
  author={Song, Chan Hee and Wu, Jiaman and Washington, Clayton and Sadler, Brian M and Chao, Wei-Lun and Su, Yu},
  booktitle={Proceedings of the IEEE/CVF international conference on computer vision},
  pages={2998--3009},
  year={2023}
}

@inproceedings{kim2023context,
  title={Context-aware planning and environment-aware memory for instruction following embodied agents},
  author={Kim, Byeonghwi and Kim, Jinyeon and Kim, Yuyeong and Min, Cheolhong and Choi, Jonghyun},
  booktitle={Proceedings of the IEEE/CVF International Conference on Computer Vision},
  pages={10936--10946},
  year={2023}
}

@article{lin2025flowplan,
  title={FlowPlan: Zero-Shot Task Planning with LLM Flow Engineering for Robotic Instruction Following},
  author={Lin, Zijun and Tang, Chao and Ye, Hanjing and Zhang, Hong},
  journal={arXiv preprint arXiv:2503.02698},
  year={2025}
}

@misc{vlmap,
      title={Visual Language Maps for Robot Navigation}, 
      author={Chenguang Huang and Oier Mees and Andy Zeng and Wolfram Burgard},
      year={2023},
      eprint={2210.05714},
      archivePrefix={arXiv},
      primaryClass={cs.RO},
      url={https://arxiv.org/abs/2210.05714}, 
}

@misc{procthor,
      title={ProcTHOR: Large-Scale Embodied AI Using Procedural Generation}, 
      author={Matt Deitke and Eli VanderBilt and Alvaro Herrasti and Luca Weihs and Jordi Salvador and Kiana Ehsani and Winson Han and Eric Kolve and Ali Farhadi and Aniruddha Kembhavi and Roozbeh Mottaghi},
      year={2022},
      eprint={2206.06994},
      archivePrefix={arXiv},
      primaryClass={cs.AI},
      url={https://arxiv.org/abs/2206.06994}, 
}

@misc{ai2thor,
      title={AI2-THOR: An Interactive 3D Environment for Visual AI}, 
      author={Eric Kolve and Roozbeh Mottaghi and Winson Han and Eli VanderBilt and Luca Weihs and Alvaro Herrasti and Matt Deitke and Kiana Ehsani and Daniel Gordon and Yuke Zhu and Aniruddha Kembhavi and Abhinav Gupta and Ali Farhadi},
      year={2022},
      eprint={1712.05474},
      archivePrefix={arXiv},
      primaryClass={cs.CV},
      url={https://arxiv.org/abs/1712.05474}, 
}

@article{batra2020rearrangement,
  title={Rearrangement: A challenge for embodied ai},
  author={Batra, Dhruv and Chang, Angel X and Chernova, Sonia and Davison, Andrew J and Deng, Jia and Koltun, Vladlen and Levine, Sergey and Malik, Jitendra and Mordatch, Igor and Mottaghi, Roozbeh and others},
  journal={arXiv preprint arXiv:2011.01975},
  year={2020}
}

@inproceedings{gan2022threedworld,
  title={The threedworld transport challenge: A visually guided task-and-motion planning benchmark towards physically realistic embodied ai},
  author={Gan, Chuang and Zhou, Siyuan and Schwartz, Jeremy and Alter, Seth and Bhandwaldar, Abhishek and Gutfreund, Dan and Yamins, Daniel LK and DiCarlo, James J and McDermott, Josh and Torralba, Antonio and others},
  booktitle={2022 International conference on robotics and automation (ICRA)},
  pages={8847--8854},
  year={2022},
  organization={IEEE}
}

@inproceedings{anderson2018vision,
  title={Vision-and-language navigation: Interpreting visually-grounded navigation instructions in real environments},
  author={Anderson, Peter and Wu, Qi and Teney, Damien and Bruce, Jake and Johnson, Mark and S{\"u}nderhauf, Niko and Reid, Ian and Gould, Stephen and Van Den Hengel, Anton},
  booktitle={Proceedings of the IEEE conference on computer vision and pattern recognition},
  pages={3674--3683},
  year={2018}
}

@inproceedings{qi2020reverie,
  title={Reverie: Remote embodied visual referring expression in real indoor environments},
  author={Qi, Yuankai and Wu, Qi and Anderson, Peter and Wang, Xin and Wang, William Yang and Shen, Chunhua and Hengel, Anton van den},
  booktitle={Proceedings of the IEEE/CVF Conference on Computer Vision and Pattern Recognition},
  pages={9982--9991},
  year={2020}
}

@misc{behavior1k,
      title={BEHAVIOR-1K: A Human-Centered, Embodied AI Benchmark with 1,000 Everyday Activities and Realistic Simulation}, 
      author={Chengshu Li and Ruohan Zhang and Josiah Wong and Cem Gokmen and Sanjana Srivastava and Roberto Martín-Martín and Chen Wang and Gabrael Levine and Wensi Ai and Benjamin Martinez and Hang Yin and Michael Lingelbach and Minjune Hwang and Ayano Hiranaka and Sujay Garlanka and Arman Aydin and Sharon Lee and Jiankai Sun and Mona Anvari and Manasi Sharma and Dhruva Bansal and Samuel Hunter and Kyu-Young Kim and Alan Lou and Caleb R Matthews and Ivan Villa-Renteria and Jerry Huayang Tang and Claire Tang and Fei Xia and Yunzhu Li and Silvio Savarese and Hyowon Gweon and C. Karen Liu and Jiajun Wu and Li Fei-Fei},
      year={2024},
      eprint={2403.09227},
      archivePrefix={arXiv},
      primaryClass={cs.RO},
      url={https://arxiv.org/abs/2403.09227}, 
}

@misc{goatbench,
      title={GOAT-Bench: A Benchmark for Multi-Modal Lifelong Navigation}, 
      author={Mukul Khanna and Ram Ramrakhya and Gunjan Chhablani and Sriram Yenamandra and Theophile Gervet and Matthew Chang and Zsolt Kira and Devendra Singh Chaplot and Dhruv Batra and Roozbeh Mottaghi},
      year={2024},
      eprint={2404.06609},
      archivePrefix={arXiv},
      primaryClass={cs.AI},
      url={https://arxiv.org/abs/2404.06609}, 
}

@misc{gpt5,
      title={OpenAI GPT-5 System Card}, 
      year={2025},
      eprint={2601.03267},
      archivePrefix={arXiv},
      primaryClass={cs.CL},
      url={https://arxiv.org/abs/2601.03267}, 
}

\clearpage
\appendix

\setcounter{page}{1}
\setcounter{equation}{0}
\renewcommand{\theequation}{\thesection\arabic{equation}}

{\LARGE \textbf{Appendix}}\\[1em]

\section{Comparison with Existing Embodied Benchmarks}

Table~\ref{tab:vln_benchmarks} compares LongAct with representative embodied benchmarks in terms of task horizon, interactivity, and goal specification.
Compared with classical navigation benchmarks such as R2R and REVERIE, LongAct requires substantially longer trajectories, increasing the average number of required steps from around 8 to more than 500.
Among existing benchmarks, GOAT and Behavior-1k are the closest to LongAct in task horizon, with average lengths of about 200 and 150 steps, respectively, but each misses one key property.
GOAT remains a navigation-only benchmark without physical interaction, while Behavior-1k supports interaction but does not use free-form task specifications.

Specifically, Behavior-1k defines 1,000 task types through PDDL-style symbolic goals.
Although its 2025 challenge releases natural-language instructions for 50 task categories, these are annotations for a subset of predefined task types rather than open-vocabulary user requests.
In contrast, LongAct combines all three properties: very long-horizon execution, physical interaction that changes object states, and free-form natural-language instructions.
This makes LongAct a more demanding testbed for embodied agents that must jointly perform long-term planning, grounded interaction, and free-form instruction following.

\begin{table}[h]
\centering
\small
\setlength{\tabcolsep}{3pt}   % 默认6pt，调小让列更紧凑
\renewcommand{\arraystretch}{1.1}

\caption{\textbf{Comparison to embodied benchmarks.}
We compare task horizon, interactivity, and goal specification.
“Interactive” indicates whether the agent performs physical interactions that change object states in the scene.
“Free-form Inst.” indicates whether task objectives are provided as open-vocabulary natural-language requests, rather than predefined task categories or structured goal predicates. For Behavior-1k, the statistics are computed from human demonstration data provided in the Behavior-1k Challenge 2025. For benchmarks marked with *, the step count is measured by the number of high-level decisions rather than low-level action steps.}
\label{tab:vln_benchmarks}

\begin{tabular}{
l
>{\centering\arraybackslash}p{1.0cm}
>{\centering\arraybackslash}p{1.0cm}
>{\centering\arraybackslash}p{1.0cm}
>{\centering\arraybackslash}p{1.0cm}
>{\centering\arraybackslash}p{1.0cm}
>{\centering\arraybackslash}p{1.0cm}
>{\centering\arraybackslash}p{1.2cm}
}
\toprule
 & \rotatebox{24}{R2R*} 
 & \rotatebox{24}{REVERIE*} 
 & \rotatebox{24}{GOAT} 
 & \rotatebox{24}{ALFRED} 
 & \rotatebox{24}{VirtualHome*} 
 & \rotatebox{24}{Behavior-1k} 
 & \rotatebox{24}{\textbf{LongAct}} \\
\midrule
Avg. Steps & $8$ & $8$ & $200$ & 50 & 11 & 150 & \textbf{$>500$} \\
Interactive & \ding{55} & \ding{55} & \ding{55} & \ding{51} & \ding{51} & \ding{51} & \textbf{\ding{51}} \\
Free-form Inst.  & \ding{51} & \ding{51} & \ding{51} & \ding{51} & \ding{51} & \ding{55} & \ding{51}\\
\bottomrule
\end{tabular}

\end{table}

\section{Failure Analysis}

\begin{figure}[h]
    \centering
    \includegraphics[width=\linewidth]{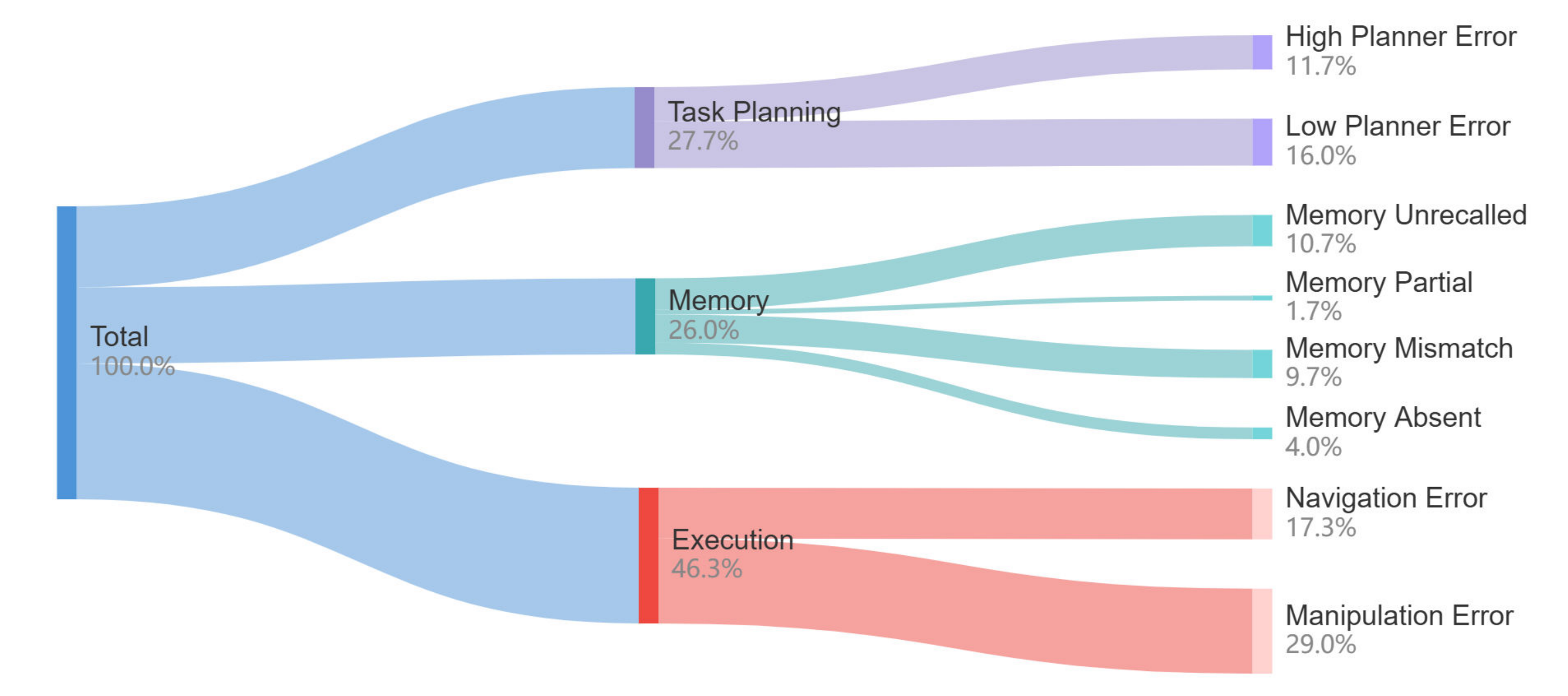}
    \caption{
    \textbf{Error distribution across task planning, memory, and execution.} This highlights that the most embodied forms of interaction—fine-grained manipulation—remain the most challenging aspect for an LLM-based agent.
    }
    \label{fig:error}
\end{figure}

We conduct an error analysis over the 301 failed items in the final-state checklists and group the errors into three major categories: \textit{Task Planning errors}, which arise from incorrect decompositions or goal sequencing produced by HoloMind’s two-layer Planner; \textit{Memory errors}, which occur either because an object was never discovered or because it existed in memory but led to failure due to being located incorrectly, retrieved wrongly, or retrieved incompletely; and \textit{Execution errors}, which result from navigation failing to reach the required position or from incorrect manipulation actions. As shown in Fig.~\ref{fig:error}, Execution is the most challenging component, and manipulation is the hardest aspect within it, indicating that for an LLM-based agent, the most embodied forms of interaction remain the most difficult to master.

\section{Demonstrations}

\begin{figure*}[h]
    \centering
    \includegraphics[width=\linewidth]{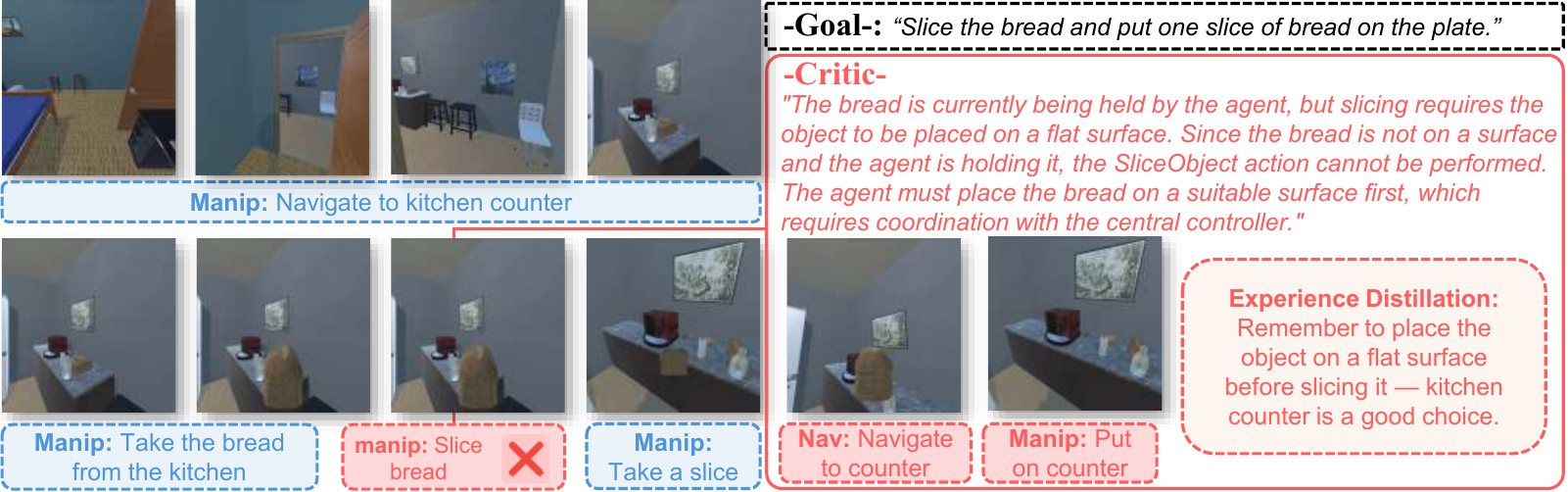}
    \caption{
        \textbf{Demonstration of critic-assisted correction during execution.}
        When the agent fails to execute \emph{Slice bread}, the Critic inspects the failure, identifies that the bread is not placed on a surface, and proposes a corrective action while updating its experience.
    }
    \label{fig:b}
\end{figure*}

\begin{figure}[h]
    \centering
    \includegraphics[width=0.9\linewidth]{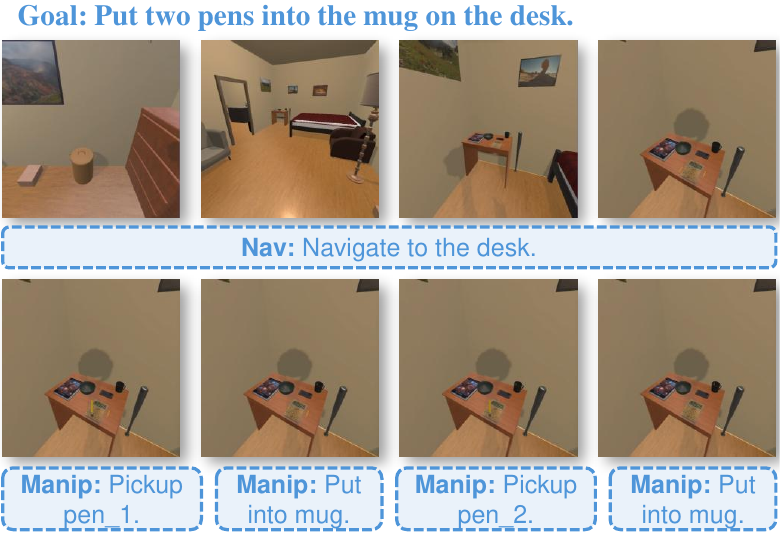}
    \caption{
        \textbf{Demonstration of sequential object handling with persistent identity tracking.}
        HoloMind is able to pick up one pen, correctly place it, and then pick up the other, maintaining a stable understanding that the two pens are distinct objects across the entire sequence of actions.
    }
    \label{fig:c}
\end{figure}

\begin{figure}[h]
    \centering
    \includegraphics[width=0.9\linewidth]{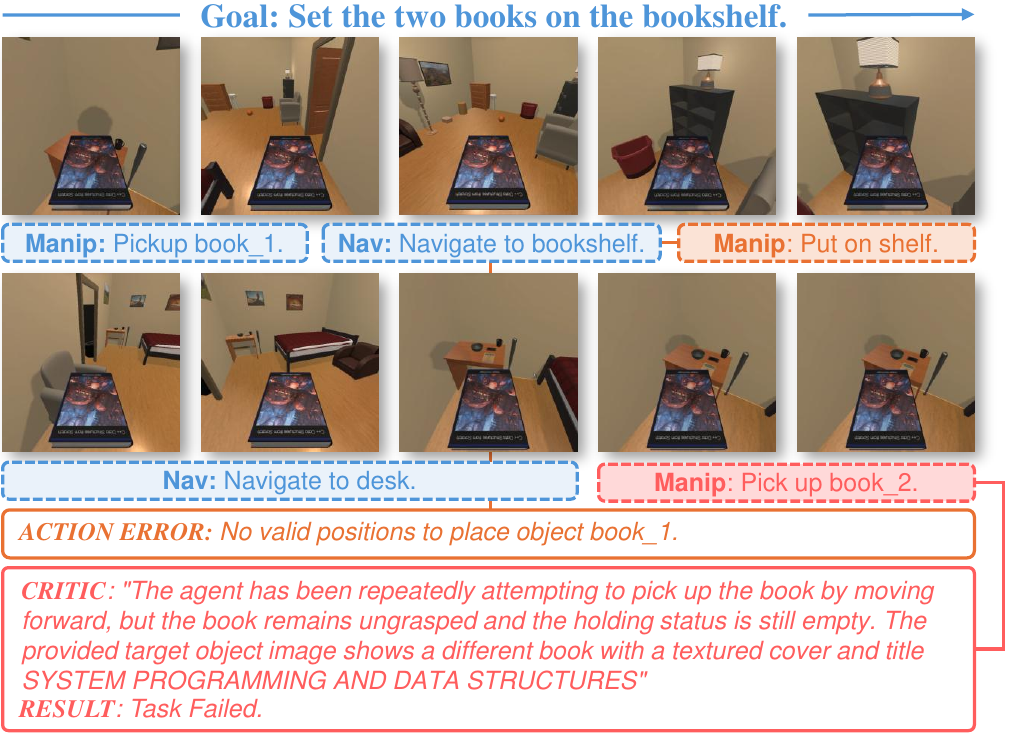}
    \caption{
        \textbf{Demonstration of a failure case with unresolved errors.}
        The agent incorrectly places the first book but fails to detect the mistake. When attempting to pick up the second book, the system becomes inconsistent, the Critic cannot recover, and control is returned to the planner, terminating the task.
    }
    \label{fig:d}
\end{figure}

\begin{figure}[t]
    \centering
    \includegraphics[width=0.9\linewidth]{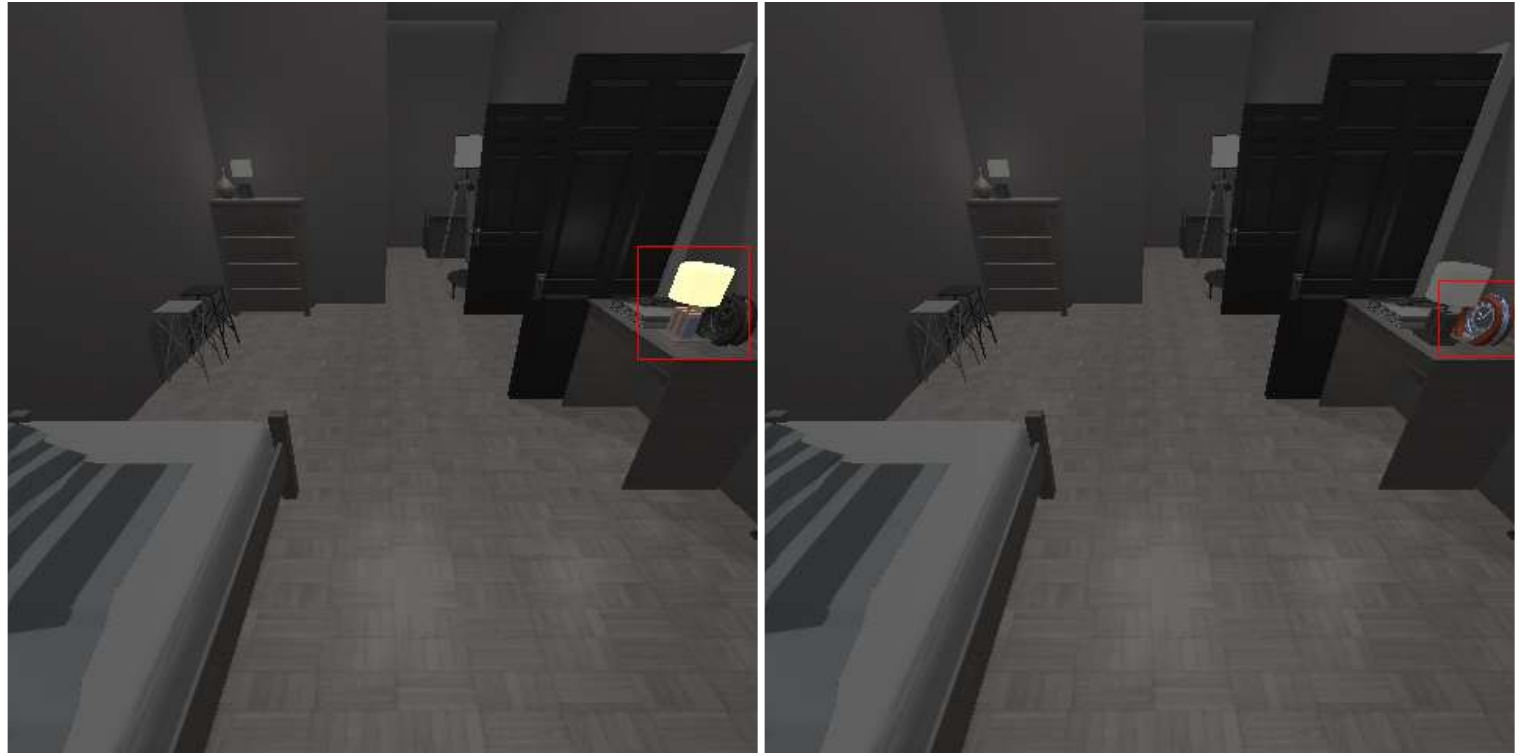}
    \caption{
        \textbf{Multimodal evidence used for memory retrieval.}
        The LLM receives visual candidates from CLIP and determines whether each image matches the query description, enabling disambiguation in cluttered or visually similar scenes.
    }
    \label{fig:llmimg}
\end{figure}

Figures~\ref{fig:b}, \ref{fig:c}, and \ref{fig:d} illustrate concrete execution trajectories.  
In Fig.~\ref{fig:b}, when the agent cannot perform \emph{Slice bread}, the Critic diagnoses the failure, identifies that the bread is not placed on a supporting surface, and recommends a correction while recording the experience.  
In Fig.~\ref{fig:c}, HoloMind demonstrates sequential object handling: it picks up one pen, places it, and then picks up the second pen, maintaining persistent identity tracking of both objects throughout the sequence.  
Fig.~\ref{fig:d} shows a failure case: the agent incorrectly places the first book but fails to notice the mistake. When it later tries to pick up the second book, inconsistencies emerge; the Critic cannot recover and ultimately returns control to the planner, causing the task to terminate.

Fig.~\ref{fig:llmimg} visualizes the multimodal information used during memory retrieval. Given CLIP-selected candidates, the LLM evaluates whether each image satisfies the query description, resolving spatial and semantic ambiguities during execution.

\section{Details of Improvement Rate Metric}

We have introduced a new metric, \textit{Improvement Rate} (\textbf{IR}). In this section, we provide a detailed discussion of why such a metric is important and why IR takes its current mathematical form.

\subsection{Motivation of IR}
The central challenge of long-horizon embodied tasks lies not only in whether an agent can eventually complete a goal, but also in whether it can \textbf{continuously and reliably improve its own performance during execution}. In realistic application scenarios, even though a highly capable assistant from the very beginning would be ideal, this is extremely difficult to achieve with current technology. A more practical and equally important expectation is that an agent—despite a potentially modest initial performance—should become more competent through ongoing interaction with the environment, gradually improving its efficiency through repeated trial-and-error, feedback, and experience accumulation. Such ``getting better over time’’ behavior is often more indicative of long-term utility than a single static measure of success.

Moreover, as the capabilities of foundation models continue to advance, we have good reason to expect that the overall learning and adaptation potential of embodied agents will improve accordingly. However, a key open question remains: \textbf{when placed in concrete environments and tasked with concrete objectives, do current agents actually exhibit signs of self-improvement?} That is, does their performance trend improve as execution unfolds? Existing static metrics cannot answer this question. They can only reveal ``how well’’ the agent ultimately performs, but not ``whether it becomes better while acting.’’

This is precisely the motivation for introducing \textit{Improvement Rate}. We require a metric that captures the agent’s \textbf{learning dynamics}—one that quantifies how the \textit{rate of change} in its score evolves over time, rather than merely evaluating the agent through a single final-outcome measure. The goal of IR is to answer one of the most fundamental questions in embodied intelligence:

\begin{quote}
\textit{As the agent progresses through a task, does it exhibit increasingly efficient behavior?}
\end{quote}

To address this, we design IR with a two-level structure:

\begin{itemize}
    \item \textbf{Local trends:} The score sequence is partitioned into multiple contiguous segments under different segmentation granularities. Within each segment, we compute the local growth rate, enabling the metric to capture short-term behavioral patterns.
    \item \textbf{Trend of trends:} These local growth rates are then linearly regressed over segment order, measuring whether the growth rate itself increases as execution proceeds.
\end{itemize}

This hierarchical design makes IR a principled indicator of an agent’s \textbf{self-improvement dynamics}. Unlike conventional static success metrics, IR explicitly targets the agent’s ability to ``get better within the same task,’’ a capability that is essential for long-horizon embodied intelligence but remains invisible to traditional evaluation protocols.

Having outlined the motivation for IR and the key intuition behind its design, we now provide a formal definition of the metric. 

\subsection{Formal Definition of IR}
\paragraph{Notation and Basic Objects.}

Let:

\begin{itemize}
  \item \(T\) denote the total task length (number of steps);

  \item the score sequence be 
    \begin{equation}
      \{s_t\}_{t=0}^{T},
    \end{equation}

  \item \(\mathcal{L}(\cdot)\) denote the operator that performs ordinary-least-squares (OLS) linear regression and returns the fitted slope. 
        Specifically, for a sequence \(X=\{x_i\}_{i=1}^{m}\), let
        \begin{equation}
          (\alpha^*, \beta^*) 
          = \arg\min_{\alpha,\beta} \sum_{i=1}^{m} \left(x_i - (\alpha\, i + \beta)\right)^2,
        \end{equation}
        and define
        \begin{equation}
          \mathcal{L}(X) = \alpha^*.
        \end{equation}
\end{itemize}

Here, \(T\) may differ across tasks or trajectories, and the operator \(\mathcal{L}\) provides a unified notion of ``local improvement rate'' under linear approximation.

\paragraph{Multi-resolution Segmentation.}

To avoid relying on a single time scale, we segment the sequence at multiple levels.
For each segmentation number 
\begin{equation}
  k \in \{2,3,\dots,n\},
\end{equation}
we partition the sequence uniformly into \(k\) contiguous intervals:
\begin{equation}
  S_{k,j} = \bigl\{ s_t \mid t \in I_{k,j} \bigr\}, \qquad j = 1,\dots,k.
\end{equation}

These intervals reflect different temporal resolutions:
\begin{itemize}
  \item Small \(k\): coarse resolution, capturing the overall trend.
  \item Large \(k\): fine resolution, sensitive to local changes.
\end{itemize}

\paragraph{First Layer: Local Growth Rates.}

Within each interval, we compute the local slope:
\begin{equation}
  \alpha_{k,j} = \mathcal{L}(S_{k,j}), \qquad j = 1,\dots,k.
\end{equation}

\paragraph{Second Layer: Slope of Slopes.}

To capture whether the agent is improving faster over time, we treat
\begin{equation}
  \{\alpha_{k,1},\alpha_{k,2},\dots,\alpha_{k,k}\}
\end{equation}
as a sequence indexed by \(j\), and compute:
\begin{equation}
  a_k = \mathcal{L}\bigl(\{\alpha_{k,1},\dots,\alpha_{k,k}\}\bigr).
\end{equation}

If \(a_k>0\), the later segments exhibit higher growth rates—i.e., the agent’s improvement is accelerating.  
If \(a_k\approx0\), the growth rate is largely steady.  
If \(a_k<0\), the agent’s improvement is slowing down.

\paragraph{Multi-scale Averaging: Final Definition of IR.}

To ensure robustness across time scales, we average over all segmentation levels:
\begin{equation}
  I = \frac{1}{n-1} \sum_{k=2}^{n} a_k.
\end{equation}

Thus \(I\), the Improvement Rate (IR), is a scalar: larger values indicate stronger acceleration in performance improvement.

\subsection{Design Details and Rationale}

In this subsection we explain the key design choices behind IR.

\paragraph{Why Multi-resolution Segmentation?}

IR relies on segmenting the score sequence at multiple resolutions rather than fixing a single segmentation level.
Using only one segmentation number \(k\) is problematic for several reasons.

First, different agents may exhibit improvement at very different temporal scales. Some models quickly adapt and then plateau, while others remain flat for a long period before experiencing a sudden breakthrough. A fixed segmentation may either smooth out such events (if the segments are too long) or fragment them into pieces that are indistinguishable from noise (if the segments are too short).

Second, because \(\{s_t\}\) is an \emph{accumulated} score sequence, extremely fine partitions (where each segment contains only a handful of steps) tend to yield almost flat segments whose fitted slopes are numerically close to zero: most of the time the cumulative score stays constant and changes only occasionally. Such ultra-short segments therefore provide little information about the underlying learning dynamics.

By considering multiple segmentation numbers \(k \in \{2,\dots,n\}\), IR captures both coarse and moderately fine temporal structure without relying on any single, arbitrary choice of time scale. This multi-resolution design allows IR to reflect:

\begin{itemize}
  \item early-phase behaviour (e.g., prolonged exploration),
  \item mid-phase improvement (e.g., discovering a more efficient strategy),
  \item late-phase patterns (e.g., saturation or gradual refinement),
\end{itemize}

\paragraph{Why Compute the Slope \(a_k\) of the Slopes \(\alpha_{k,j}\)?}

While \(\alpha_{k,j}\) captures the local growth rate within each segment, the quantity that truly reflects \emph{learning efficiency} is how these local slopes themselves evolve over time. By fitting the sequence
\begin{equation}
  \{\alpha_{k,1},\alpha_{k,2},\dots,\alpha_{k,k}\}
\end{equation}
with a linear model and taking its slope \(a_k\), we obtain a scalar that summarises:

\begin{quote}
\emph{Is the agent’s scoring growth rate increasing, stable, or decreasing over the course of the task at segmentation level \(k\)?}
\end{quote}

If \(a_k > 0\), later segments exhibit higher growth rates, indicating that the agent is improving faster as the episode progresses. If \(a_k \approx 0\), the agent’s improvement rate is roughly steady. If \(a_k < 0\), the agent’s ability to gain additional score is fading over time.

Aggregating these quantities across different segmentation levels then yields the final IR metric, which captures this ``trend of trends’’ in a way that is robust to the particular temporal scale at which self-improvement manifests.

% ---------------------- Figure 1 ----------------------
\begin{figure}[h]
    \centering
    \begin{minipage}[t]{0.45\linewidth}
        \centering
        \includegraphics[width=\linewidth]{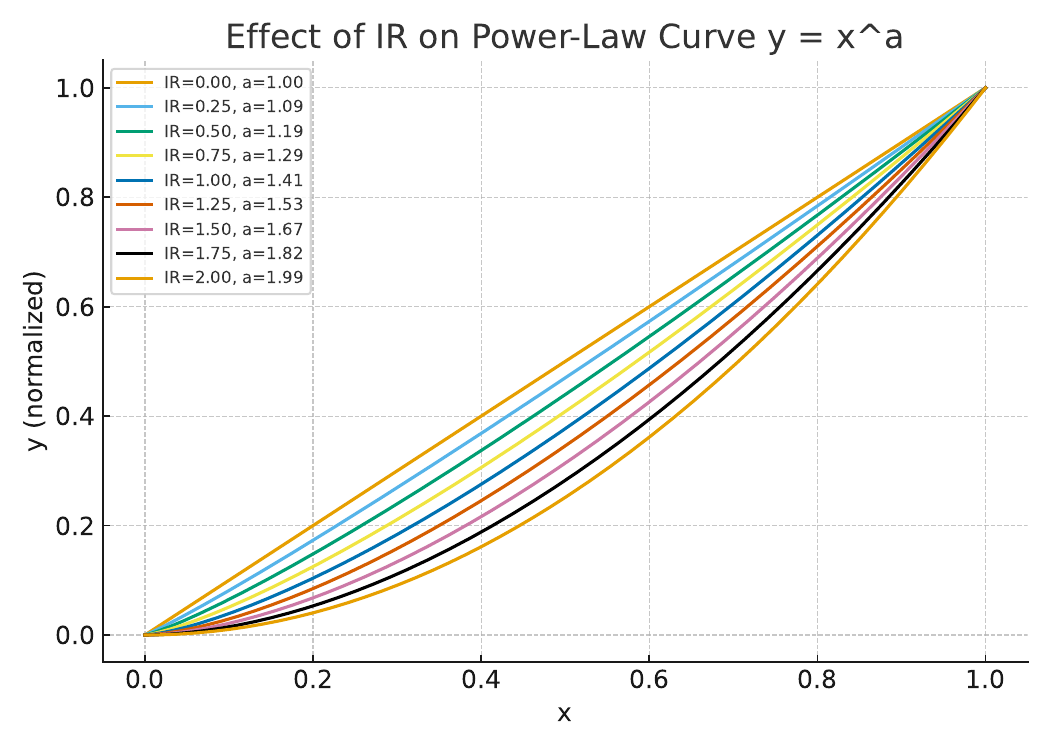}
        \caption{
            \textbf{Power-law curves for IR values from $0$ to $2$.}
            We numerically determine the exponent $a$ such that $s_t = (t/T)^a$ produces the desired IR value, 
            yielding a continuous sweep of geometric shapes.
        }
        \label{fig:ir_sweep}
    \end{minipage}
    \hfill
    \begin{minipage}[t]{0.45\linewidth}
        \centering
        \includegraphics[width=\linewidth]{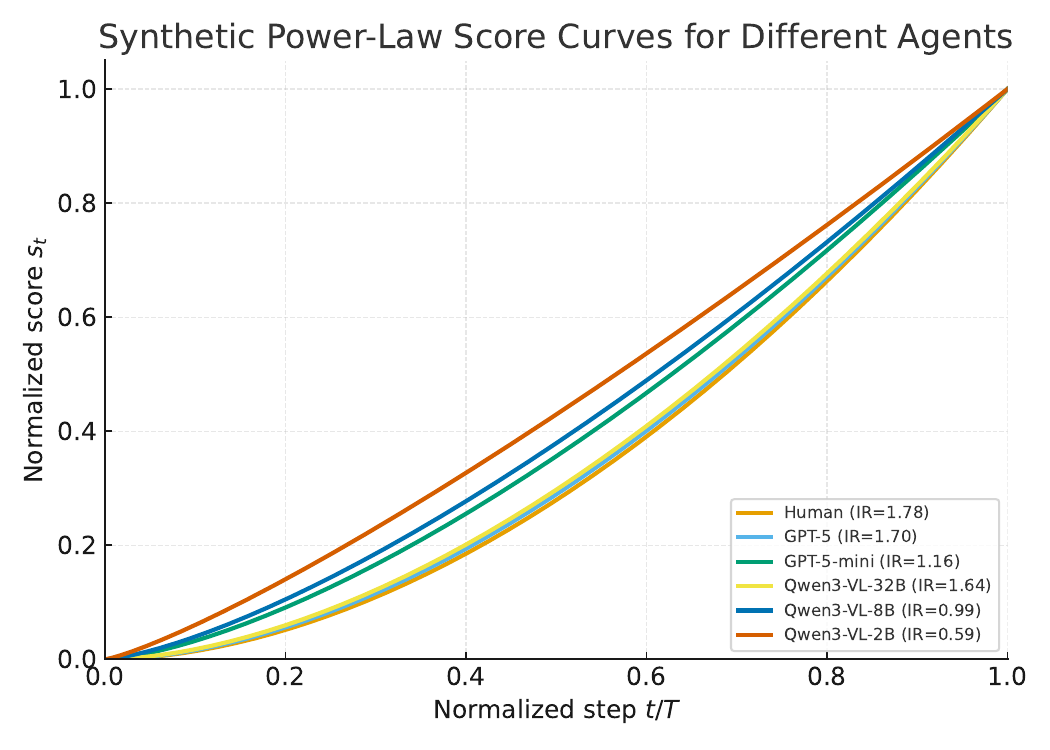}
        \caption{
            \textbf{Comparison of synthetic curves corresponding to different agents.}
            Each agent’s IR is mapped to a power-law curve $s_t = (t/T)^a$ with the matched exponent $a$, 
            enabling a consistent geometric comparison of self-improvement tendencies.
        }
        \label{fig:ir_real_models}
    \end{minipage}
\end{figure}

\subsection{Visualization and Geometric Interpretation}
To better illustrate the meaning of the Improvement Rate (IR) and how it captures different levels of self-improvement, we provide two complementary visualizations based on synthetic power-law curves of the form $s_t = (t/T)^a$. Although these curves are not actual trajectories, they offer a clear geometric interpretation of how IR relates to the 
shape of improvement over time.

\paragraph{IR Sweep: How Curve Shape Varies with IR.}
As shown in Figure~\ref{fig:ir_sweep}, sweeping IR from $0$ to $2$ yields a smooth progression of synthetic curves that 
illustrate the effect of IR on curvature. Low IR values correspond to nearly linear shapes, while high IR values produce 
increasingly convex profiles that reflect stronger acceleration in improvement. This figure serves as a canonical reference 
for understanding IR in geometric terms.

\paragraph{Model Comparison: Placing Agents into the IR Geometry.}
Figure~\ref{fig:ir_real_models} applies the same synthetic construction to the agents evaluated in our benchmark, 
including humans, GPT-5, GPT-5-mini, and the Qwen3-VL family. By mapping each measured IR value to its corresponding 
power-law curve, we obtain a unified geometric view of how rapidly different agents would improve under idealized growth 
patterns. The relative curvature among these curves offers an intuitive comparison of their self-improvement tendencies.

% \section{}

\end{document}